\documentclass[sigconf]{acmart}
\AtBeginDocument{%
  }

\copyrightyear{2026}
\acmYear{2026}
\setcopyright{cc}
\setcctype{by}
\acmConference[KDD '26]{Proceedings of the 32nd ACM SIGKDD Conference on Knowledge Discovery and Data Mining V.1}{August 09--13, 2026}{Jeju Island, Republic of Korea}
\acmBooktitle{Proceedings of the 32nd ACM SIGKDD Conference on Knowledge Discovery and Data Mining V.1 (KDD '26), August 09--13, 2026, Jeju Island, Republic of Korea}
\acmPrice{}
\acmDOI{10.1145/3770854.3780317}
\acmISBN{979-8-4007-2258-5/2026/08}

\usepackage{graphicx}
\usepackage{booktabs}
\usepackage[table]{xcolor}
\usepackage{stfloats}

\begin{document}

\title{CoPHo: Classifier-guided Conditional Topology Generation with Persistent Homology}


\author{Gongli Xi}
\affiliation{%
  \institution{School of Cyberspace Security, Beijing University of Posts and Telecommunications}
  \city{Beijing}
  \country{China}
}
\email{kevinxgl@bupt.edu.cn}

\author{Ye Tian}
\authornote{Corresponding author: Ye Tian.}
\affiliation{%
  \institution{State Key Laboratory of Networking and Switching Technology, Beijing University of Posts and Telecommunications}
  \city{Beijing}
  \country{China}  
}
\email{yetian@bupt.edu.cn}

\author{Mengyu Yang}
\affiliation{%
  \institution{State Key Laboratory of Networking and Switching Technology, Beijing University of Posts and Telecommunications}
  \city{Beijing}
  \country{China}
}
\email{mengyuyang@bupt.edu.cn}

\author{Zhenyu Zhao}
\affiliation{%
  \institution{State Key Laboratory of Networking and Switching Technology, Beijing University of Posts and Telecommunications}
  \city{Beijing}
  \country{China}
}
\email{1781065115@bupt.edu.cn}

\author{Yuchao Zhang}
\affiliation{%
  \institution{School of Computer Science (National Pilot Software Engineering School), Beijing University of Posts and Telecommunications}
  \city{Beijing}
  \country{China}
}
\email{yczhang@bupt.edu.cn}

\author{Xiangyang Gong}
\affiliation{%
  \institution{State Key Laboratory of Networking and Switching Technology, Beijing University of Posts and Telecommunications}
  \city{Beijing}
  \country{China}
}
\email{xygong@bupt.edu.cn}

\author{Xirong Que}
\affiliation{%
  \institution{State Key Laboratory of Networking and Switching Technology, Beijing University of Posts and Telecommunications}
  \city{Beijing}
  \country{China}
}
\email{rongqx@bupt.edu.cn}

\author{Wendong Wang}
\affiliation{%
  \institution{State Key Laboratory of Networking and Switching Technology, Beijing University of Posts and Telecommunications}
  \city{Beijing}
  \country{China}
}
\email{wdwang@bupt.edu.cn}
\renewcommand{\shortauthors}{Gongli Xi et al.}

\begin{abstract}
The structure of topology underpins much of the research on performance and robustness, yet available topology data are typically scarce, necessitating the generation of synthetic graphs with desired properties for testing or release. Prior diffusion-based approaches either embed conditions into the diffusion model, requiring retraining for each attribute and hindering real-time applicability, or use classifier-based guidance post-training, which does not account for topology scale and practical constraints. In this paper, we show from a discrete perspective that gradients from a pre-trained graph‐level classifier can be incorporated into the discrete reverse diffusion posterior to steer generation toward specified structural properties. Based on this insight, we propose \textbf{C}lassifier-guided \textbf{Co}nditional Topology Generation with \textbf{P}ersistent \textbf{Ho}mology  (\textbf{CoPHo}), which builds a persistent homology filtration over intermediate graphs and interprets features as guidance signals that steer generation toward the desired properties at each denoising step. Experiments on four generic/network datasets demonstrate that CoPHo outperforms existing methods at matching target metrics, and we further validate its transferability on the QM9 molecular dataset. The code is available at \url{https://github.com/Lrbomchz/CoPHo}.
\end{abstract}



\maketitle

\section{Introduction}
Topology lies at the core of many applications (e.g., communication networks and protocol design) \cite{leskovec2010signed,leskovec2007graph}. The arrangement of nodes and links fundamentally impacts routing efficiency, latency, throughput, fault tolerance, and other performance metrics. Prior studies have shown that protocol behavior (e.g. multicast scaling laws, routing state overhead) can vary dramatically across different topological configurations \cite{Bai2019}. As a result, network designers must carefully consider topology in optimizing network performance and reliability.
\begin{figure}[htbp]        
  \centering               
  \includegraphics[width=0.45\textwidth]{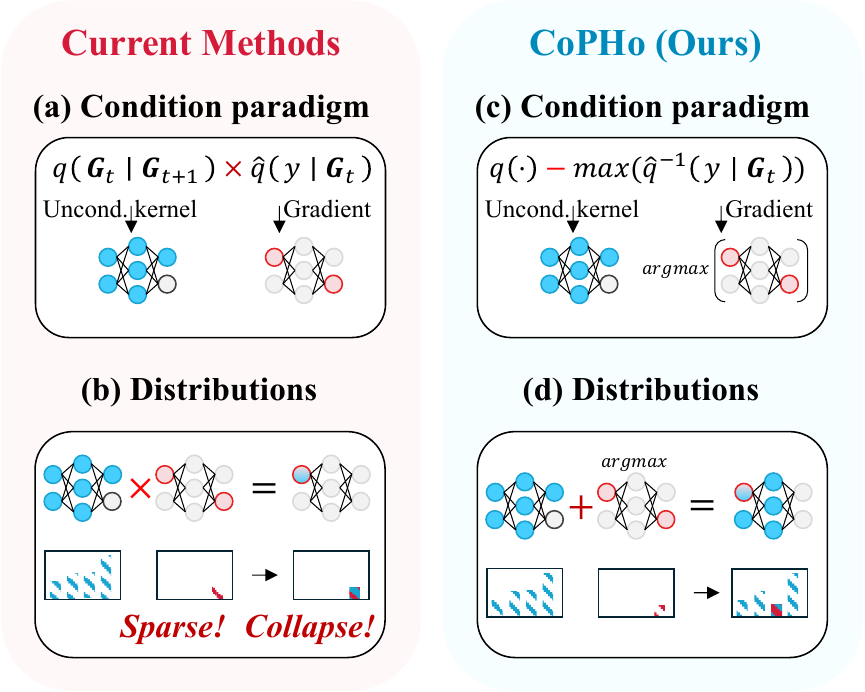}
  \caption{Comparison between the current methods and CoPHo. (a), (c) provide a high-level explanation of how CoPHo differs from existing methods, while (b) and (d) show that sparse gradients on continuous graphs cause collapse; CoPHo instead discretely edits the graph to avoid this.}
  \label{fig:moti}
  \Description{motivation}
\end{figure}
However, obtaining and experimenting with real network topologies is often difficult. Carrier and enterprise network graphs are typically proprietary, and even research testbeds reveal only limited or anonymized structural information due to security and privacy concerns \cite{qian2017social}. This motivates the use of synthetic graph topology generators for network planning, simulation, benchmarking, and data sharing \cite{alrumaih2023genind,dadauto2024data}. Indeed, realistic synthetic topologies serve as important tools to evaluate new protocols and algorithms in lieu of real networks. Such tools allow researchers to simulate network behavior on large-scale, representative topologies and to perform “what-if” analyses for new designs. Synthetic topology generation also facilitates anonymization: one can release a plausible network graph for public use without disclosing the true network, as long as the synthetic graph preserves key structural patterns of the original \cite{yang2020secure,yuan2023privgraph}. In general, the ability to algorithmically generate graphs with specified characteristics is critical for both networking research and practical network engineering. This motivates the problem of \emph{conditional topology generation}: \textbf{\textit{automatically creating graphs that satisfy user-defined constraints.}}

Early network topology generators were largely rule-based or random \cite{tangmunarunkit2002network,sharafaldin2019developing,gutfraind2015multiscale}. These approaches have limited ability to capture complex structural patterns observed in real-world networks. In the past decade, data-driven approaches \cite{You2018,ermagun2021recovery,patro2012missing} have emerged to learn generative models of graphs \cite{You2018}. Deep graph generative models such as variational autoencoders \cite{Simonovsky2018} and generative adversarial networks \cite{DeCao2018} have been applied to produce graphs that emulate training examples. More recently, diffusion-based generative models have achieved remarkable success in image and audio synthesis \cite{kong2020diffwave,rombach2022latent,popov2021gradtts}, prompting their adoption for graph data. Diffusion models generate data by iteratively denoising random noise, effectively learning the reverse of a gradual noising process. Several researchers \cite{chen2023efficient,gebauer2019symmetry,gebauer2022inverse,geng2023novo,gruver2023protein,li2023graphmaker,nisonoff2024unlocking,Jo2022,Vignac2023DiGress,jiang2024diffkg,liu2024graphdif,you2024latent,liu2024graph,sharma2024diffuse} have extended this paradigm to graph data, demonstrating its promise for graph generation.

Despite these advances, significant challenges remain in controlling the topologies generated by diffusion models, which is an essential requirement for practical use in network topology design. Existing conditional diffusion–based approaches introduce constraints either during the forward noising process \cite{Mercatali2024} or by applying classifier guidance in the reverse sampling phase \cite{Lee2023}. Methods that integrate conditions into the forward stochastic differential equation train a distinct conditional score network for each target attribute, but this strategy imposes heavy computational burdens and requires extensive training data for every new specification, rendering it impractical when confronted with unseen conditions. 
Current classifier guidance methods \cite{Dhariwal2021DiffusionBeatGANs}, by contrast, apply classifier gradients at every denoising step . However, real-world network graphs are highly sensitive to even minor link changes, small edge perturbations can fracture connectivity or undermine resilience \cite{molnar2023threshold,cavallaro2024sensitivity}. 
Moreover, practical topologies tend to be extremely sparse (edge density below 10\%) \cite{melancon2006just,casiraghi2025empirical}, and training data are limited. Consequently, practitioners favor message-passing GNNs, such as GCN \cite{kipf2016semi}, GraphSAGE \cite{hamilton2017inductive}, and GIN \cite{xu2018powerful}, whose cost scales with the number of edges \(O(E)\), rather than fully-connected architectures \cite{kreuzer2021rethinking,yun2019graph,ying2021transformers,kong2023goat} with \(O(n^2)\) complexity. In these message‐passing GNNs, gradients only propagate along existing edges; as shown in Fig.~\ref{fig:moti}, the sparsity of the gradient signals can cause the generated graph to collapse.

To address the limitations of existing methods by unifying classifier guidance and topological optimization under the diffusion paradigm, we propose \textbf{C}lassifier-guided \textbf{Co}nditional Topology Generation with \textbf{P}ersistent \textbf{Ho}mology (CoPHo). In theory, we first establish that the gradient of a pretrained graph-property predictor can be rigorously integrated into the reverse-time denoising steps to enforce constraints without retraining the base diffusion model. Building on this foundation, CoPHo treats conditioning as an optimization problem in the space of persistent homology, leveraging the formal correspondence between classifier gradients and topological feature distances.

In practice, CoPHo proceeds by computing at each denoising step the gradient of the classifier with respect to the current unconditioned graph and interpreting its magnitude as a point-wise distance measure, and then constructs a persistent homology filtration \cite{Edelsbrunner2002} over the sequence of intermediate graphs. This filtration framework naturally aligns with the edge-localized gradient flow of message-passing graph neural networks by modeling an initial contraction of nonessential links followed by guided expansion to satisfy the target properties,  and can be extended to fully connected architectures.

Finally, CoPHo leverages persistent homology to guide classifier-based edits in an adaptive schedule, enabling conditioning over target properties at each diffusion step.  Ablation studies on diverse network datasets show that CoPHo consistently outperforms baseline conditional diffusion models in matching specified properties and maintaining robust local structure.  Additional trials on molecular graph benchmarks confirm that this topology-aware guidance generalizes effectively across graph domains.

In summary, our contributions are summarized as follows:
\begin{itemize}
\item (\textbf{Conceptual \& Methodological})  
Adopting a fully discrete perspective, we model each reverse‐time update as a Markov chain.  We show that gradients from a graph property classifier can directly drive these discrete node/edge edits to satisfy the desired properties at every diffusion step.
\item(\textbf{Technical})  
We incorporate persistent homology into the discrete diffusion process by applying a filtration on each denoised graph to extract persistence scores, then use these scores to guide node/edge edits in stages. This structured and topology‐aware scheduling counteracts the sparsity‐induced under-connectivity common in GNNs.
\item (\textbf{Empirical}) Extensive evaluation on Planar, Enzymes, Com-munity-small and Topology Zoo network benchmarks, as well as molecular graphs, demonstrates superior fidelity to target degree distributions, clustering coefficients and diameter specifications, while preserving robust local connectivity and validating cross-domain transferability.
\end{itemize}

\section{Related Work}
\paragraph{Graph Diffusion and Training with Conditional Inputs}
In conditional diffusion training, the target properties are embedded into the denoising network so that each reverse step predicts the previous graph conditioned on desired attributes \cite{Mercatali2024,xu2024discrete}. Conditions are encoded and concatenated with graph and timestep representations, enabling end‐to‐end multi‐property generation \cite{hoogeboom2022equivariant,huang2024learning,xu2023geometric}. Each time a new property or property combination is required, the entire model must be retrained or extensively fine‐tuned.
\paragraph{Classifier‐based Conditioning in Graph Diffusion}
Existing methods decouple diffusion training from conditioning by first learning an unconditional graph diffusion model and then deriving gradient‐based corrections from an external property predictor \cite{bao2022equivariant, Vignac2023DiGress}.  Extensions like MOOD further incorporate out‐of‐distribution guidance signals during reverse sampling \cite{Lee2023}.  However, by treating these gradient signals as generic continuous perturbations, they overlook (i) the inherent message‐passing constraints of GNN architectures—where gradients flow only along existing edges—and (ii) the discrete, combinatorial nature of graph topology. 
\paragraph{Persistent Homology in GNNs}
A few recent works have explored persistent homology (PH) to preserve topological features in GNN models \cite{hofer2020graph}.  PH‐induced graph ensembles use Vietoris–Rips filtrations on time‐series sensor networks to construct multiple graph views, but focus on representation rather than conditional generation \cite{nguyen2025persistent}. Besides, PH summaries have been used in GNN pooling layers to retain homological invariants in classification tasks, yet these methods do not address graph synthesis \cite{yan2025enhancing}.  To our knowledge, CoPHo is the first to integrate PH into the denoising \emph{and} conditioning loop with both theoretical justification and scalable implementation.

\section{Method}
\subsection{Preliminaries}
\label{sec:prelim}

\paragraph{Diffusion Models.}
Existing diffusion models fall into two main architectures: the Denoising Diffusion Probabilistic Model (DDPM) and the Score‐based Stochastic Differential Equation (SDE) framework.  
In the DDPM framework, the forward noising process is defined as a discrete‐time Markov chain  
\begin{equation}
q(\mathbf{x}_{t+1}\mid\mathbf{x}_t)
=\mathcal{N}\bigl(\mathbf{x}_{t+1};\,\sqrt{1-\beta_t}\,\mathbf{x}_t,\;\beta_t\,\mathbf{I}\bigr),
\tag{1.1}
\end{equation}
and the learned reverse denoising model takes the form  
\begin{equation}
p_\theta(\mathbf{x}_t\mid\mathbf{x}_{t+1})
=\mathcal{N}\bigl(\mathbf{x}_t;\,\mu_\theta(\mathbf{x}_{t+1},t),\;\Sigma_\theta(\mathbf{x}_{t+1},t)\bigr),
\tag{1.2}
\end{equation}
which can be trained by predicting the noise \(\epsilon_\theta(\mathbf{x}_{t+1},t)\) \cite{Ho2020DDPM,Song2019Score}.

In contrast, the continuous‐time SDE formulation \cite{Song2020ScoreSDE} treats the forward process as  
\begin{equation}
d\mathbf{x} = f(\mathbf{x},t)\,dt + g(t)\,d\mathbf{w},
\tag{2.1}
\end{equation}
where \(f\) and \(g\) specify the drift and diffusion coefficients and \(\mathbf{w}\) is a Wiener process.  Its reverse‐time dynamics are governed by  
\begin{equation}
d\mathbf{x} = \bigl[f(\mathbf{x},t) - g(t)^2\nabla_{\mathbf{x}}\log p_t(\mathbf{x})\bigr]\,dt + g(t)\,d\bar{\mathbf{w}},
\tag{2.2}
\end{equation}
which explicitly constructs a vector field \(f - g^2\nabla\log p_t\) that transports samples back to regions of high density.  Under this SDE perspective, classifier gradients can be added directly as extra drift terms at selected time points, enabling even sparse guidance to effectively steer generation.  


\paragraph{Discrete Graph Diffusion.}
Let a graph be \(\mathbf{G}_t=(\mathbf{V},\mathbf{E}_t)\), where \(\lvert \mathbf{V}\rvert=n\).  Node attributes lie in a discrete set \(\mathcal{X}\) of size \(a\), and edge attributes in \(\mathcal{E}\) of size \(b\).  Define forward transition matrices \(\mathbf{Q}^V_t\in[0,1]^{a\times a}\) and \(\mathbf{Q}^E_t\in[0,1]^{b\times b}\).  The forward noising is
\begin{equation}
q(\mathbf{V}_{t+1}\mid \mathbf{V}_t)
=\mathrm{Cat}\bigl(\mathbf{V}_{t+1};\,\mathbf{V}_t\,\mathbf{Q}^V_t\bigr),
q(\mathbf{E}_{t+1}\mid \mathbf{E}_t)
=\mathrm{Cat}\bigl(\mathbf{E}_{t+1};\,\mathbf{E}_t\,\mathbf{Q}^E_t\bigr)
\tag{3}
\end{equation}
and the reverse model \(p_\theta(\mathbf{G}_t\mid\mathbf{G}_{t+1})\) is learned via a GNN to approximate the true posterior \cite{Vignac2023DiGress}\cite{Qin2023SparseDiff}.

\paragraph{Persistent Homology (PH).}
Given a weighted graph \(\mathbf{G}=(\mathbf{V},\mathbf{E},W)\), define a \emph{decreasing} filtration \(\{\mathcal{F}_\alpha\}\) by
\begin{equation}
\mathcal{F}_\alpha
=\bigl(\mathbf{V},\,\{\,e\in\mathbf{E}:W(e)\le\alpha\}\bigr)
\tag{4}
\end{equation}
so that as \(\alpha\) decreases, edges are monotonically removed.  Persistent homology tracks the \emph{birth} \(b_i\) and \emph{death} \(d_i\) of homological features, producing a persistence diagram \(\{(b_i,d_i)\}\) whose lifetimes \(d_i-b_i\) measure feature significance \cite{Zomorodian2005,Edelsbrunner2008PH}.  These diagrams yield differentiable topological summaries for gradient-based graph optimization.
\paragraph{Motivation for Integrating Persistent Homology.}
\textbf{First}, message‐passing GNNs inherently propagate gradients only along existing edges, leading to monotonic edge removals during classifier‐guided updates; this behavior aligns intuitively with the decreasing filtration in persistent homology, and the associated edge‐weight ordering naturally extends to fully‐connected or multi‐type GNN architectures.  
\textbf{Second}, direct gradient injections on sparse graphs can cause structural collapse (see Fig.~\ref{fig:moti}), as even small continuous perturbations may sever critical connections.  By filtering edges through a homology‐based threshold, CoPHo enforces a controlled, multi‐scale simplification that preserves global connectivity while steering toward desired properties.  
\paragraph{Classifier-Based Conditioning.} 
According to the derivation of \citet{Dhariwal2021DiffusionBeatGANs}, let \(\hat{q}(\,\cdot\mid \mathbf{y}_G)\) denote the conditional denoising kernel and \(q(\,\cdot)\) the unconditional one.  In the ideal conditional diffusion model, at each reverse step one samples
\begin{equation}
\label{eq:ori_cond}
\hat q(\mathbf{x}_t \mid \mathbf{x}_{t+1}, \mathbf{y})\,
\tag{5}
\end{equation}
This sampling procedure can be shown to satisfy
\begin{equation}
\label{eq:ori_decomp}
\hat q(\mathbf{x}_t \mid \mathbf{x}_{t+1}, \mathbf{y})
=\frac{q(\mathbf{x}_t \mid \mathbf{x}_{t+1})\,\hat q(\mathbf{y}\mid \mathbf{x}_t)}{\hat q(\mathbf{y}\mid \mathbf{x}_{t+1})}\,
\tag{6}
\end{equation}

Here \(\hat q(\mathbf{y}\mid \mathbf{x}_{t+1})\) is independent of \(\mathbf{x}_t\) and can be treated as a constant.  The term \(q(\mathbf{x}_t\mid \mathbf{x}_{t+1})\) is approximated by a pretrained denoising network \(p_\theta(\mathbf{x}_t\mid \mathbf{x}_{t+1})\), while \(\hat q(\mathbf{y}\mid \mathbf{x}_t)\) can be estimated by an auxiliary classifier \(\Phi\).

DiGress \cite{Vignac2023DiGress} extends this framework to graphs by treating \(\mathbf{G}\) as a continuous variable.  However, for discrete graph properties \(\mathbf{y}\), such as the length of the unique shortest \(\mathrm{path} = |\{e_1,\dots,e_n\}|\) between two nodes, \(\mathbf{G}\) is not differentiable.  If one removes an edge \(\mathbf{e}_\delta\), then:
\[
\begin{cases}
e_\delta\notin\{e_1,\dots,e_n\}\;\Longrightarrow\;\mathrm{path}=n,\\
e_\delta\in\{e_1,\dots,e_n\}\;\Longrightarrow\;\mathrm{path}\text{ jumps to the next‐best path.}
\end{cases}
\]
Consequently, injecting classifier gradients directly into the unconditional denoising process yields only suboptimal conditioning.
\paragraph{Key Notation.} The key symbols used by CoPHo are listed in Table~\ref{tab:notation}.
\begin{table}[ht]
  \centering
  \caption{Key Notation and Definitions}
  \label{tab:notation}
  \begin{tabular}{@{}ll@{}}
    \toprule
    \textbf{Symbol} & \textbf{Meaning} \\
    \midrule
    $t$ & Diffusion timestep index. \\
    $\mathbf{G}_t$ & Graph at step $t$, with nodes $\mathbf{V_t}$ and edges $\mathbf{E}_t$. \\
    $\mathbf{\hat G}_t$ & Conditioned graph at denoising step $t$. \\
    $q(\cdot)$ & Unconditional denoising kernel. \\
    $\hat q(\cdot)$ & Conditional denoising kernel. \\
    $y$ & Desired target property (e.g., clustering coefficient). \\
    $\Phi$ & Auxiliary classifier/regressor for predicting $y$. \\
    $T_{\mathrm{homo}}$ & Number of PH steps per diffusion update. \\
    $\mathbf{\hat G}_t^i$ & Candidate graph at diffusion step $t$ and PH step $i$. \\
    $\varphi(\mathbf{\hat G}_t^i)$ & True property value of candidate graph $\mathbf{\hat G}_t^i$. \\
    $w(\mathbf{\hat G}_t^i)$ & Importance weight for candidate graph $\mathbf{\hat G}_t^i$. \\
    $\mathbb{P}(\mathbf{\hat G}_t\mid \mathbf{G}_t)$ & Proposal distribution for importance sampling. \\
    $g_t(e),\,g_t(v)$ & Edge/node gradient scores w.r.t. classifier ($\Phi$) loss. \\
    $\mathcal{F}_\alpha$ & Decreasing filtration at threshold $\alpha$. \\
    $\alpha_i$ & Threshold at homology step $i$. \\
    \bottomrule
  \end{tabular}
\end{table}

\subsection{Guided Proposals in Graph Diffusion – A Discrete View}
\label{sec:proof_brief}
\begin{figure*}[ht]        
  \centering               
  \includegraphics[width=0.9\textwidth]{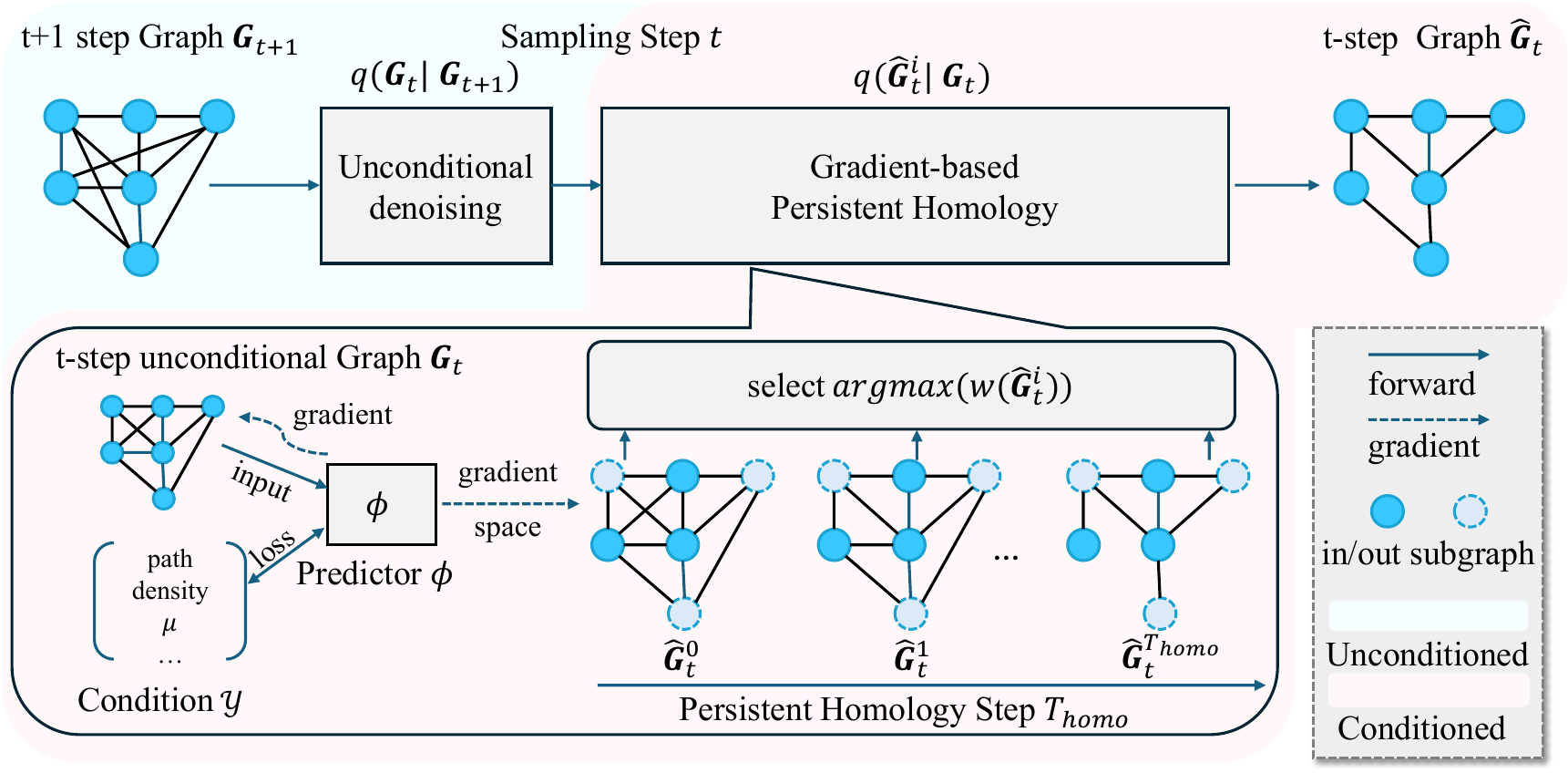}
  \caption{Overview of \textbf{CoPHo}. At each denoising step, we build a decreasing filtration via persistent homology to capture multi-scale topological features. A local subgraph around the conditioned nodes and the full graph are fed into a lightweight GNN \(\Phi\), which outputs node-level and edge-level gradient signals. These signals drive monotonic node and edge removals, producing a sequence of subgraphs that converges to a final graph \(\widehat G\) meeting the desired global and fine-grained properties without retraining the diffusion backbone.}
    \Description{Diagram showing the CoPHo framework: persistent homology filtration, GNN processing, and monotonic node/edge removals leading to a final graph.}
  \label{fig:model_overview}
\end{figure*}

We reframe conditional graph generation without relying on continuity. As shown in Figure~\ref{fig:model_overview}, let \(\mathbf{G}_{t+1}\) be the noisy graph at step \(t+1\), \(\mathbf{G}_t\) its unconditional denoising samples, and \(\mathbf{\hat G}_t\) the desired conditional sample.  We define each sampling step of the conditional diffusion as a two‐stage Markov chain:
\begin{equation}
\label{eq:graph_markov}
\mathbf{G}_{t+1}\;\longrightarrow\;\mathbf{G}_t\;\longrightarrow\;\mathbf{\hat G}_t
\tag{7}
\end{equation}

From Equation~\ref{eq:graph_markov}, we model the sampling process from \(\mathbf{G}_{t+1}\) to \(\mathbf{\hat G}_t\).  By the law of total probability, the chain rule, and the Markov property (derive in Appendix Section~\ref{sec:derivation_sampling}), we obtain:
\begin{equation}
\label{eq:full_prob}
q(\mathbf{\hat G}_t\mid \mathbf{G}_{t+1})
=\sum_{\mathbf{G}_t}q(\mathbf{\hat G}_t\mid \mathbf{G}_t)\;q(\mathbf{G}_t\mid \mathbf{G}_{t+1})\,.
\tag{8}
\end{equation}
The key term is 
\(\,q(\mathbf{\hat G}_t\mid \mathbf{G}_t)\),
which we decompose via Bayes’ rule by introducing the condition \(y\):
\begin{equation}
\label{eq:markov_bas}
q(\mathbf{\hat G}_t\mid \mathbf{G}_t)
=\frac{q(y\mid \mathbf{G}_t)}{q(y\mid \mathbf{\hat G}_t)}\;q(\mathbf{\hat G}_t\mid \mathbf{G}_t,y)\,
\tag{9}
\end{equation}
Here \(q(y\mid \mathbf{G}_t)\) and \(q(y\mid\mathbf{\hat G}_t)\) are the probabilities that \(\mathbf{G}_t\) and \(\mathbf{\hat G}_t\) satisfy \(y\).  Since \(\mathbf{G}_t\) arises from the unconditional process, \(q(y\mid \mathbf{G}_t)\) is independent of \(\mathbf{G}_t\)’s generation and can be treated as constant; \(\mathbf{\hat G}_t\) by construction satisfies \(y\).  Thus our goal reduces to modeling 
\(\,q(\mathbf{\hat G}_t\mid \mathbf{G}_t,y)\).

Directly sampling from the conditional kernel is intractable due to the noncontinuous nature of \(\mathbf G\).  Instead, we adopt an importance‐sampling scheme: we use a tractable proposal distribution \(\mathbb P(\hat{\mathbf G}_t\mid \mathbf G_t)\), draw \(N\) (\(N=T_{homo}\) after introducing PH) candidates \(\{\mathbf{\hat G}_t^i\}_{i=1}^N\), and retain those satisfying the condition \(y\). In the Appendix~\ref{sec:proof_gradient_proposal} we prove that using the gradient of a predictor \(\Phi(\mathbf G)\) as a surrogate for \(q(y\mid \mathbf{G})\) yields a valid and effective proposal $\mathbb P$.  Empirical results in Table~\ref{tab:ablation_space} confirm the practical superiority of classifier‐gradient proposals over other sampling. For each candidate \(\mathbf{\hat G}_t^i\), to ensure unbiased estimation we assign the importance weight (see proofs in  Appendix~\ref{sec:proof_unbiased_weight}):
\begin{equation}
w(\mathbf{\hat G}_t^i)
=\frac{q(\mathbf{\hat G}_t^i \mid \mathbf{G}_{t+1})\;q(y \mid \mathbf{\hat G}_t^i)}
      {\mathbb P(\mathbf{\hat G}_t^i \mid \mathbf{G}_t)}\,
\tag{10}
\end{equation}

We decompose \(w\) into two interpretable factors:
\textbf{1). Condition probability:}  
   \(
     q\bigl(y \mid \mathbf{\hat G}_t^i\bigr)
   \)
   is the likelihood that \(\mathbf{\hat G}_t^i\) satisfies the desired property \(y\).  In practice this is computed directly from the property’s definition (e.g.\ shortest‐path length, clustering coefficient) on \(\mathbf{\hat G}_t^i\).
\textbf{2). Transition ratio:}
   \(
     \frac{q(\mathbf{\hat G}_t^i \mid \mathbf{G}_{t+1})}
          {\mathbb P(\mathbf{\hat G}_t^i \mid \mathbf{G}_t)}
   \)
is inversely proportional to the edit distance \(\| \mathbf{G}_t - \mathbf{\hat G}_t^i\|\). In the model, this means that while \(\mathbf{\hat G}_t^i\) must satisfy the target property \(y\), we seek to minimally modify the unconditional diffusion sample \(\mathbf{G}_t\).

\subsection{Classifier-guided Discrete Conditional Generation with Persistent Homology}
Building on the theoretical foundation established in Section~\ref{sec:proof_brief}, we now address the practical challenges of sparse graph data. In this section, we leverage message‐passing GNNs to encode graph edits as a monotonic  persistent homology filtration: at each diffusion step, we apply a homology‐driven removal of features in descending order of persistence, ensuring that the evolving graph both remains realistic and increasingly satisfies \(y\). Note that, as discussed in Section~\ref{conclus}, our framework extends to both fully‐connected GNNs (e.g., graph transformers) and to multi‐class node/edge‐type GNNs.  In our QM9 experiments, we employ a fully‐connected Graph Transformer backbone and model atoms and bonds as multi‐class features.  Table~\ref{tab:qm9} demonstrates CoPHo’s transferability to these settings.
\paragraph{Integration into the Persistent Homology Framework}
Given a weighted graph \(\mathbf{G}=(\mathbf{V},\mathbf{E},W)\), we first define the usual \emph{decreasing} filtration
\begin{equation}
\mathbf{\mathcal{F}}_\alpha
=\bigl(\{v\in\mathbf{V}:W(v)\le\alpha\},\{\,e\in\mathbf{E}:W(e)\le\alpha\}\bigr)
\tag{11}
\end{equation}
so that as \(\alpha\) decreases, edges are removed in order of increasing weight.  To incorporate our gradient‐guided conditioning, we modify this filtration at each denoising step \(t\) as follows:

1. \textbf{Gradient‐based subgraph selection.}  Let \(\mathbf{G}_{t}\) be the current noisy graph.  We compute per‐node and edge gradient scores 
\begin{equation}
g_t(e)\;=\;\nabla_e \mathrm{CE}\bigl(\Phi(\mathbf{G}_{t}),y_G\bigr),\;
g_t(v)\;=\;\nabla_v \mathrm{CE}\bigl(\Phi(\mathbf{G}_{t}),y_G\bigr)
\tag{12}
\end{equation}
then sort nodes and edges by descending \(\,g_t\) and \(\,g_t\).  We select the top-\(k\) nodes and edges (those with the most positive gradients determined by \(T_{homo}\)) to form the initial subcomplex \(\mathbf{S}_t^0\subseteq\mathbf{\mathcal{F}}_{\alpha_{\max}}=\mathbf{G}_t\).

2. \textbf{Monotonic simplification.}  Starting from \(\mathbf{S}_t^0\), we introduce a fixed number of homology steps \(T_{homo}\).  At step \(i\in\{1,\dots,T_{homo}\}\), we choose a threshold \(\alpha_i\) halfway between the \((k_i)\)th and \((k_{i+1})\)th largest \(g_t\) values, and define (note that \(\mathbf{\hat G}_t^i=\mathbf{G}_t  \lor \mathbf{S}_t^i\)):
\begin{equation}
\mathbf{S}_t^i=\mathbf{\mathcal{F}}_{\alpha_{i}}
=\bigl(\{\,v\in \mathbf{S}_t^{i-1}:\,g_t(v)\le\alpha_i\},\{\,e\in \mathbf{S}_t^{i-1}:\,g_t(e)\le\alpha_i\}\bigr)
\tag{13}
\end{equation}
thereby removing exactly one edge/node at each homology step, and \(i=\|\mathbf{S}_t^{i}-\mathbf{S}_t^{0}\|=\| \mathbf{G}_t - \mathbf{\hat G}_t^i\|\).

3. \textbf{Property evaluation and weighting.}  For each candidate \(\mathbf{\hat G}_t^i\), we compute the target property \(\varphi\bigl(\mathbf{\hat G}_t^i\bigr)\) (e.g.\ clustering coefficient or shortest‐path length).  To bias the diffusion update toward satisfying \(y_G\) while preserving the unconditional dynamics, we form weight for each \(\mathbf{\hat G}_t^i\):

\begin{equation}
\label{eq:weights}
w\bigl(\mathbf{\hat G}_t^i\bigr)
\sim
\underbrace{e^{-\|\mathbf{\hat G}_t^i-\mathbf{ G}_t\|}}_{\sim\;\displaystyle\frac{q(\hat{\mathbf G}_t^i\mid\mathbf G_{t+1})}
    {\mathbb P(\hat{\mathbf G}_t^i\mid\mathbf G_t)}}
\;\times\;
\underbrace{\bigl(\varphi(\mathbf{\hat G}_t^i)-y\bigr)^{-2}}_{\sim\;\displaystyle q\bigl(y\mid\hat{\mathbf G}_t^i\bigr)}
\;\times\;
\mathbf{1}\!\Bigl[\bigl|\varphi(\mathbf{\hat G}_t^i)-y\bigr|\le\epsilon\Bigr]
\tag{14}
\end{equation}

\begin{table*}[htbp]
\centering
\caption{Results (↓) for single‐global‐property conditioning on Community‐Small.}
\label{tab:global_single}
\resizebox{\linewidth}{!}{
\begin{tabular}{lcccc|cccc}
\toprule
 & \multicolumn{4}{c}{Community-small} & \multicolumn{4}{c}{Enzymes} \\
 \midrule
Model     & Density & Clustering & Assortativity & Transitivity 
          & Density & Clustering & Assortativity & Transitivity \\
\midrule
GDSS      & 2.95 & 12.1 & 19.6 & 11.4 
          & 8.04 & 2.53 & 1.98 & 2.55 \\
GDSS-T    & 2.30 & 11.5 & 19.2 & 10.1 
          & 9.25 & 3.27 & 2.03 & 2.68 \\
DiGress   & 2.32 & 10.3 & 16.8 &  9.22 
          & 8.11 & 2.39 & 1.93 & 2.41 \\
MOOD-1    & 2.35 & 11.1 & 18.8 & 10.5 
          & 7.94 & 2.34 & 1.83 & 2.12 \\
MOOD-4    & 2.12 & 11.3 & 16.7 &  8.76 
          & 7.98 & 2.44 & 1.99 & 2.43 \\
Twigs     & \cellcolor{gray!20}2.07 &  \cellcolor{gray!20}9.67 & \cellcolor{gray!20}15.2 &  \cellcolor{gray!20}8.35 
          & \textbf{7.35} & \cellcolor{gray!20}2.23 & \cellcolor{gray!20}1.72 & \cellcolor{gray!20}2.03 \\
\midrule
\textbf{CoPHo}     & \textbf{1.89} & \textbf{7.13} & \textbf{10.9} & \textbf{7.23} 
          & \cellcolor{gray!20}7.78 & \textbf{2.21} & \textbf{1.26} & \textbf{1.75} \\
\bottomrule
\end{tabular}
}
\end{table*}

Here, 
\(\mathbf{1}[\cdot]\) denotes the indicator function, which immediately rejects any sample whose properties fall below \(\epsilon\), thereby preventing poor-quality samples.

In this way, CoPHo embeds persistent homology filtration directly into the denoising process: nodes/edges are pruned in a conditioned, monotonic sequence guided by classifier gradients, and each intermediate complex is evaluated against the desired properties.  This yields an end‐to‐end conditioning mechanism that (i) minimally perturbs the original reverse kernel, (ii) provides interpretable topological updates via the filtration \(\{\mathcal{F}_\alpha\}\), and (iii) drives the generated graph toward matching both global and fine‐grained constraints.

\section{Experiments}
\label{sec:exp}
\begin{table*}[htbp]
  \centering
  \caption{Results (↓) for multi‐global-properties conditioning on Community‐Small.}
  \label{tab:multi_control}
  \begin{tabular}{l|cc|cc|ccc}
    \toprule
              & \multicolumn{2}{c}{Pair 1} & \multicolumn{2}{c}{Pair 2} & \multicolumn{3}{c}{Triplet} \\
    Model     & assortativity & transitivity & clustering & assortativity & clustering & assortativity & transit. \\
    \midrule
    GDSS      & 19.4\textcolor{gray}{\scriptsize$\pm$0.7}           & 13.2\textcolor{gray}{\scriptsize$\pm$0.5}         & 13.1\textcolor{gray}{\scriptsize$\pm$0.7}       & 20.3\textcolor{gray}{\scriptsize$\pm$0.6}           & 13.3\textcolor{gray}{\scriptsize$\pm$0.7}       & 18.9\textcolor{gray}{\scriptsize$\pm$0.8}           & 12.8\textcolor{gray}{\scriptsize$\pm$0.5}         \\
    DiGress   & 17.5\textcolor{gray}{\scriptsize$\pm$0.6}           & 10.8\textcolor{gray}{\scriptsize$\pm$0.5}         & 11.3\textcolor{gray}{\scriptsize$\pm$0.5}       & 17.7\textcolor{gray}{\scriptsize$\pm$0.9}           & 11.8\textcolor{gray}{\scriptsize$\pm$0.5}       & 17.5\textcolor{gray}{\scriptsize$\pm$0.7}           & 10.9\textcolor{gray}{\scriptsize$\pm$0.6}         \\
    Twigs     & 16.9\textcolor{gray}{\scriptsize$\pm$0.4}           &  8.94\textcolor{gray}{\scriptsize$\pm$0.5}        & 10.5\textcolor{gray}{\scriptsize$\pm$0.3}       & 17.1\textcolor{gray}{\scriptsize$\pm$0.3}           & 10.7 \textcolor{gray}{\scriptsize$\pm$0.6}      & 15.8\textcolor{gray}{\scriptsize$\pm$0.6}           &  8.77 \textcolor{gray}{\scriptsize$\pm$0.4}       \\
    \textbf{CoPHo}     & \textbf{12.4}\textcolor{gray}{\scriptsize$\pm$0.3}           &  \textbf{7.61}\textcolor{gray}{\scriptsize$\pm$0.6}        &  \textbf{9.28 }\textcolor{gray}{\scriptsize$\pm$0.6}     & \textbf{13.8}\textcolor{gray}{\scriptsize$\pm$0.5}           &  \textbf{9.65 }\textcolor{gray}{\scriptsize$\pm$0.6}      & \textbf{11.7}\textcolor{gray}{\scriptsize$\pm$0.5}            &  \textbf{7.30 }\textcolor{gray}{\scriptsize$\pm$0.6}        \\
    \bottomrule
  \end{tabular}
\end{table*}

\subsection{Experimental Setup}
\label{sec:setup}
\paragraph{Datasets.}  

We evaluate on four graph benchmarks with varying structural complexity and an additional molecular benchmark for transfer testing. Community-small comprises \(\lvert G\rvert=200\) network topologies with clear community structure and \(\lvert N\rvert\approx 20\) nodes each (\cite{erdos1960evolution,Sen2008Collective}). Enzymes contains \(\lvert G\rvert=600\) protein tertiary-structure graphs averaging 37 nodes and 84 edges (\cite{Schomburg2004BRENDA}). The Planar set consists of \(\lvert G\rvert=500\) planar graphs with node counts \(\lvert N\rvert=64\) (\cite{Martinkus2022SPECTRE}). Topology-Zoo provides \(\lvert G\rvert=200\) real-world WAN topologies from 9 to 754 nodes (\cite{Knight2011TopologyZoo}). Finally, QM9 includes \(\sim\!133{,}885\) molecular graphs with 9–29 heavy atoms (\cite{Ramakrishnan2014QM9}). 

Community-small and Enzymes evaluate \emph{global} property conditioning—specifically density, clustering, and assortativity—because they encompass intricate global structures, such as modular community organization and functional motifs, which pose substantial challenges for conditional generation methods. Planar and Topology-Zoo test \emph{fine-grained} shortest-path conditioning, and QM9 demonstrates cross-domain generalizability.

\paragraph{Implementation Details.} 
For continuous and discrete target properties, we train separate regressors and classifiers, respectively.  On Community-small, Enzymes, Planar, and Topology-Zoo we use a six-layer message-passing GNN for property evaluation, while on QM9 we employ a Graph Transformer variant.  The diffusion models build on the \textbf{unconditional} DiGress backbone (\cite{Vignac2023DiGress}).

\paragraph{Baselines.}  
We compare to four state-of-the-art conditional diffusion frameworks. GDSS is a continuous-time score-based model that diffuses node and edge via SDEs and we employ the DiGress's classifier guidance for property conditioning (\cite{Jo2022}). DiGress implements discrete graph diffusion and steers sampling with gradients from a pretrained graph-level regressor (\cite{Vignac2023DiGress}). MOOD extends score-based diffusion with an out-of-distribution guidance mechanism, biasing reverse-time sampling toward high-property regions (\cite{Lee2023}). Diffusion Twigs introduces parallel “trunk” and “stem” diffusion flows for properties, coordinated via loop guidance to enrich conditional flexibility (\cite{Mercatali2024}).

\paragraph{Evaluation Metrics}
All experiments follow the protocol of \citet{Mercatali2024}.  For each target we perform 5 independent runs and report the mean and standard deviation.  Metrics are divided into two groups.  First, \emph{generation quality} metrics (Deg, Clus, Orb) quantify the statistical difference between generated and real graphs using Maximum Mean Discrepancy (MMD) \cite{Song2020ScoreSDE}.  Second, \emph{conditioning quality} metrics assess how closely each generated sample matches its prescribed properties via mean squared error (MSE), as in \citet{Mercatali2024}.

\begin{figure*}[htbp]        
  \centering               
\includegraphics[width=0.95\textwidth]{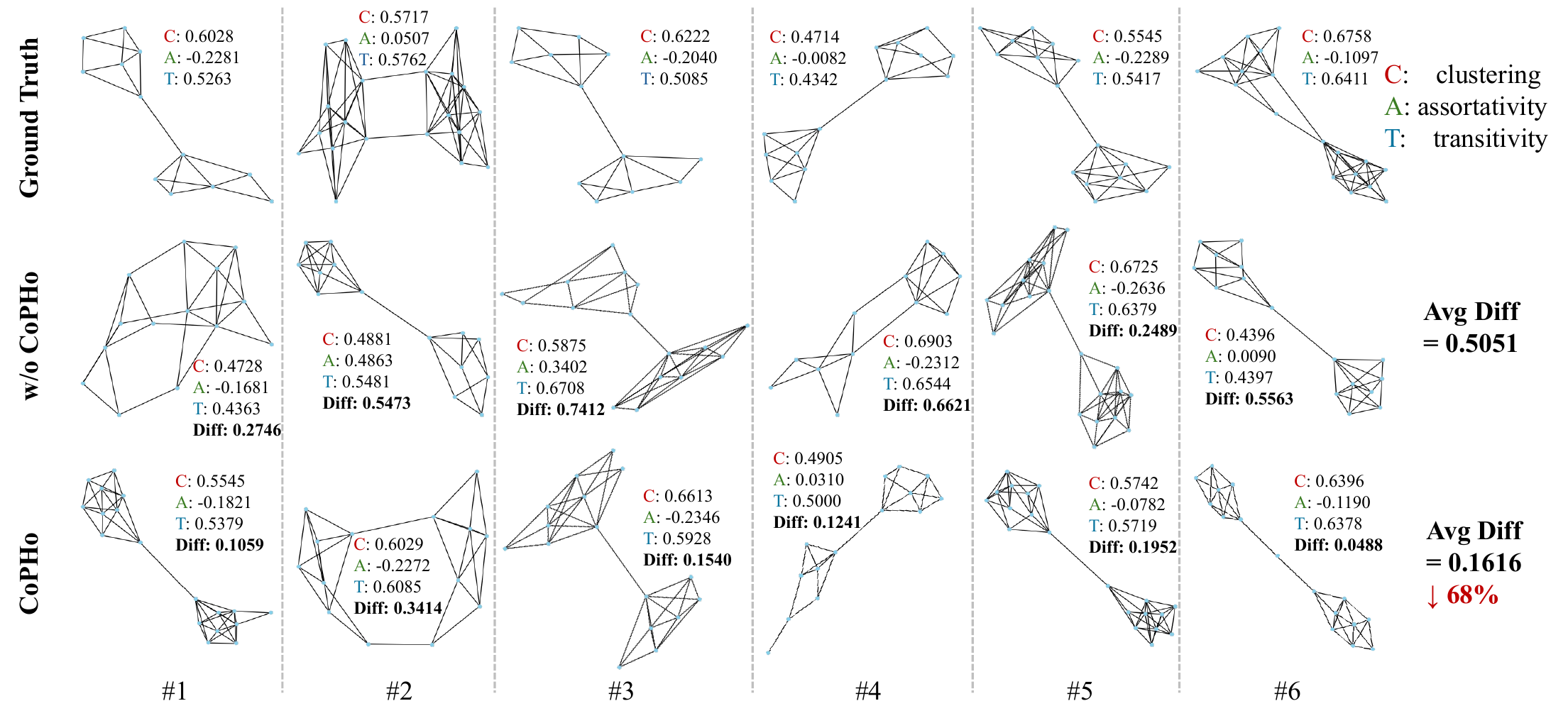}
  \caption{Visualization of Community-small. \textbf{Top}: real topologies from Community-small. Middle: The corresponding conditional samples without CoPHo. Bottom: Conditional samples with CoPHo.}
  \Description{per-sample visualization}
  \label{fig:persample}
\end{figure*}

\subsection{Global Property Conditioning}
\paragraph{Properties Extraction}
We extract four global properties with NetworkX \cite{hagberg2008exploring}. Density is defined as $\mathrm{Density} = \frac{2|E|}{|V|(|V|-1)}$. The global clustering coefficient is $C = \frac{1}{|V|}\sum_{v\in V}\frac{2\,T(v)}{k_v(k_v-1)}$, where $T(v)$ counts triangles incident to $v$ and $k_v$ is its degree. Assortativity is the Pearson correlation of degrees at the ends of each edge. Transitivity is defined as $\mathrm{Transitivity} = \frac{3\times\text{triangles}}{\text{connected triples}}$. All settings align with \citet{Mercatali2024} and \citet{Song2020ScoreSDE}.

\paragraph{Single Property Conditioning}
Table~\ref{tab:global_single} reports the mean absolute error (MAE) of three runs for Community-Small and Enzymes.  The best errors are \textbf{bolded} and second-best shaded.  CoPHo achieves the lowest MAE on clustering, assortativity, and transitivity, demonstrating superior conditional over complex global structures. 

Density conditioning leverages the fact that for fixed $|V|$, density decreases monotonically as edges are removed. Simple edge removal strategies can thus achieve a target density but they may not fully exercise the capabilities of conditional generation methods. We therefore treat density as a complementary constraint that informs but does not dominate our evaluation.

\paragraph{Multi‐properties Conditioning}
We further evaluate CoPHo for simultaneous conditioning of multiple properties.  Table~\ref{tab:multi_control} shows that CoPHo achieves the lowest MAE across all tested combinations.  To illustrate CoPHo’s per‐sample behavior under triplet conditioning, Figure~\ref{fig:persample} presents generated examples.  Samples \#1, \#2 and \#5 demonstrate that even when edge density differs from the ground truth, CoPHo preserves the target assortativity, clustering and transitivity while producing more diverse topologies.  This capability is crucial for assessing routing resilience under varied connectivity patterns.  For additional condition fusion methods and generation quality experiments, see Appendix~\ref{more_abla}.

\subsection{Fine-Grained Property Conditioning}
\label{sec:fine_control}
\begin{table}[htbp]
  \centering
  \caption{Fine-grained conditional of shortest paths, averaged over Topology-Zoo and Planar.}
  \label{tab:fine_single}
  \resizebox{\columnwidth}{!}{%
  \begin{tabular}{lccc|cc|cc}
    \toprule
                & \multicolumn{3}{c}{One Path} & \multicolumn{2}{c}{5 Paths} & \multicolumn{2}{c}{50 Paths} \\
    Model       & MAE  & KL    & OL(\%) & MAE & OL(\%) & MAE & OL(\%) \\
    \midrule
    NSP      & 0    & 0     & 100      & 0   & 100      & 0   & 100      \\
    DiGress     & 1.45\textcolor{gray}{\scriptsize$\pm$0.13} & 0.12\textcolor{gray}{\scriptsize$\pm$0.01}  & 3.17\textcolor{gray}{\scriptsize$\pm$0.02}       & 1.51\textcolor{gray}{\scriptsize$\pm$0.03}  &   \textbf{2.28}\textcolor{gray}{\scriptsize$\pm$0.01}       & 1.48\textcolor{gray}{\scriptsize$\pm$0.01}  & 2.13\textcolor{gray}{\scriptsize$\pm$0.01}         \\
    CoPHo       & \textbf{0.43}\textcolor{gray}{\scriptsize$\pm$0.01} & \textbf{0.02}\textcolor{gray}{\scriptsize $\pm$0.00}  & \textbf{3.15}\textcolor{gray}{\scriptsize$\pm$0.01}       & \textbf{1.33}\textcolor{gray}{\scriptsize$\pm$0.01}  & 2.39\textcolor{gray}{\scriptsize$\pm$0.01}         & \textbf{1.37}\textcolor{gray}{\scriptsize$\pm$0.01}  & \textbf{2.11}\textcolor{gray}{\scriptsize$\pm$0.01}         \\
    \bottomrule
  \end{tabular}%
  }
\end{table}

\paragraph{Properties Extraction.}
We target shortest paths as our fine-grained properties because controlling them requires accounting for the entire subgraph induced by all paths between each source and target pair.  We compute all-pairs shortest paths using Dijkstra’s algorithm \cite{Dijkstra1959} and train a simple multi-layer GNN to predict these path lengths.

\paragraph{Motivation for Conditioning.}
In practice one may wish to release a usable topology while preserving privacy of the true structure.  A generated graph should make the shortest paths among critical node pairs closely match those of the original graph so that routing remains accurate. A naive approach that extracts the subgraph on key nodes and adds noise (referred to as \emph{Node Subgraph Perturbation}, \textbf{NSP} in Table~\ref{tab:fine_single}, which overlaps with the original graph on key topological structures) still risks privacy breaches \cite{kim2022equalnet}.

\paragraph{Experimental Results.}
We begin by conditioning on a single shortest path and then increase the number of conditioned paths to 5 and 50. When all pairs are conditioned, the generated graph \(\widehat G\) is identical to the original \(G\).  Table~\ref{tab:fine_single} reports the mean absolute error (MAE), Kullback–Leibler divergence, and overlaping rate (OL) of the conditioned subgraph across three runs. A full discussion of generation quality effects appears in Section~\ref{gene_quality}.

\paragraph{Case Study.}
\begin{figure}[htbp]        
  \centering               
  \includegraphics[width=0.5\textwidth]{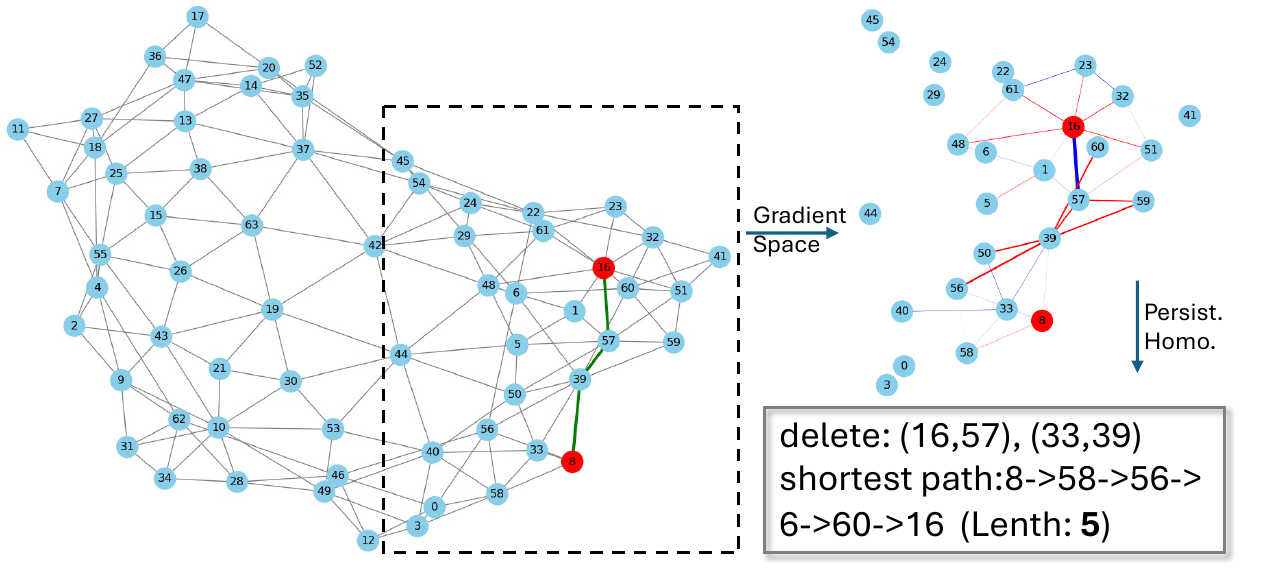}
  \caption{A case study on shortest-path conditioning. Red: negative gradient. Blue: positive gradient.}
  \label{fig:case}
  \Description{case study}
\end{figure}
To illustrate CoPHo’s fine-grained conditioning, Figure~\ref{fig:case} shows a case where we guide the shortest path between nodes \#16 and \#8.  The original distance is 3 and the target is 5.  We train a GNN to predict the path length and compute the gradient of the loss \((\Phi(\mathbf{G})-5)\) with respect to the adjacency matrix \(\mathbf{E}\).  Positive gradient entries suggest edge removal and negative entries suggest edge retention.  CoPHo introduces the path \( \#8 \to \#58 \to \#56 \to \#6 \to \#60 \to \#16 \)
and removes shorter alternatives, achieving the desired distance of 5.

\subsection{Ablation Studies}
\label{sec:ablation}

We introduce two key components in CoPHo: persistent homology and the gradient-based proposal distribution.  Our ablations cover the use of persistent homology, \(T_{homo}\), the timing of homology introduction, and the choice of proposal distribution. In this section, we report results for three factors: \(T_{\text{homo}}\), the timing of homology introduction, and different proposal distributions. Additional ablation results appear in Appendix Section~\ref{more_abla}.

\paragraph{\(T_{\text{homo}}\) and the timing of homology introduction.}
In our persistent-homology design, we introduce two key hyperparameters: the maximum number of homology steps \(T_{\text{homo}}\) and the timing of homology introduction \(\text{ph\_timing}\). The corresponding ablations are reported in Table~\ref{tab:ablation_Thomo} and Table~\ref{tab:ablation_phtiming}. 

\begin{table}[htbp]
  \centering
  \caption{Ablation on \(T_{\text{homo}}\).}
  \label{tab:ablation_Thomo}
  \begin{tabular}{cccc}
    \toprule
    \(T_{\text{homo}}\) & transitivity & assortativity & orbit \\
    \midrule
    1  & 8.5  & 13.3 & 0.088 \\
    5  & 7.2  & 10.9 & 0.092 \\
    10 & 7.9  & 8.69 & 0.092 \\
    20 & 7.7  & 14.2 & 0.112 \\
    \bottomrule
  \end{tabular}%
\end{table}

\begin{table}[htbp]
  \centering
  \caption{Ablation on the timing of homology introduction (\(\text{ph\_timing}\)).}
  \label{tab:ablation_phtiming}
  \begin{tabular}{cccc}
    \toprule
    ph\_timing & transitivity & assortativity & orbit \\
    \midrule
    0.0 & 8.4 & 11.0 & 0.102 \\
    0.2 & 7.7 & 11.2 & 0.095 \\
    0.4 & 7.8 & 10.8 & 0.092 \\
    0.6 & 7.2 & 10.9 & 0.092 \\
    0.8 & 8.9 & 14.5 & 0.089 \\
    \bottomrule
  \end{tabular}%
\end{table}

For hyperparameter selection, we balance compute and accuracy: in some cases (e.g., comparing \(T_{\text{homo}}=5\) vs.\ 10 and \(\text{ph\_timing}=0.4\) vs.\ 0.6), additional computation yields only marginal gains, so we adopt the lower-cost settings, \(T_{\text{homo}}=5\) (fewer homology steps) and \(\text{ph\_timing}=0.6\) (later PH introduction).

\begin{table}[htbp]
  \centering
  \caption{Ablation of proposal distribution on Enzymes.}
  \label{tab:ablation_space}
  \begin{tabular}{lcccc}
    \toprule
    Space                  & density & clustering & assortativity & transitivity \\
    \midrule
    Rand                   & 7.81\textcolor{gray}{\scriptsize$\pm$0.53}    & 2.35\textcolor{gray}{\scriptsize$\pm$0.06}       & 1.68\textcolor{gray}{\scriptsize$\pm$0.04}          & 2.04\textcolor{gray}{\scriptsize$\pm$0.12}         \\
    EBC & \textbf{7.59}\textcolor{gray}{\scriptsize$\pm$0.15}    & 2.36\textcolor{gray}{\scriptsize$\pm$0.06}       & 1.59\textcolor{gray}{\scriptsize$\pm$0.06}          & 1.86 \textcolor{gray}{\scriptsize$\pm$0.02}         \\
    neg EBC                & 8.27\textcolor{gray}{\scriptsize$\pm$0.32}    & 2.31\textcolor{gray}{\scriptsize$\pm$0.06}       & 1.68\textcolor{gray}{\scriptsize$\pm$0.02}          & 1.93\textcolor{gray}{\scriptsize$\pm$0.01}         \\
    Loop grad.                   & 7.80\textcolor{gray}{\scriptsize$\pm$0.41}     & 2.38\textcolor{gray}{\scriptsize$\pm$0.02}       & 1.78\textcolor{gray}{\scriptsize$\pm$0.04}          & 2.31\textcolor{gray}{\scriptsize$\pm$0.10}         \\
    Gradient               & 7.78\textcolor{gray}{\scriptsize$\pm$0.22}    & \textbf{2.21}\textcolor{gray}{\scriptsize$\pm$0.05}       & \textbf{1.26}\textcolor{gray}{\scriptsize$\pm$0.07}          & \textbf{1.75}\textcolor{gray}{\scriptsize$\pm$0.05}         \\
    \bottomrule
  \end{tabular}
\end{table}

\paragraph{Proposal Distribution}
To assess the effectiveness of gradient‐based proposals, we evaluate five proposal distribution on Enzymes and compare \textbf{rand}, which uses a random vector of the same shape as the true gradient; \textbf{EBC}, ranking edges by betweenness centrality, and \textbf{neg EBC}, its inverse ranking; \textbf{Loop gradient}, which uses the density predictor’s gradient to condition clustering, the clustering predictor’s gradient to condition assortativity, and so on; and finally \textbf{gradient}, the persistent‐homology–derived guidance at the heart of CoPHo. Table~\ref{tab:ablation_space} reports MAE over four global properties, where the gradient‐based distribution achieves the lowest error in clustering, assortativity, and transitivity, confirming its superior condition effectiveness.   

\subsection{More Results}
\label{sec:moreres}
\paragraph{QM9 Generation}
We sample 100 molecules from the QM9 test set and retrieve their dipole moment \(\mu\) and highest occupied molecular orbital (HOMO).  We then apply CoPHo to condition \(\mu\), HOMO, and both simultaneously.  To explore hybrid strategies, we combine DiGress and CoPHo in three ways: \emph{DiGress+CoPHo} uses DiGress guidance for the first 50\% of diffusion steps and CoPHo for the remaining steps, \emph{CoPHo+DiGress} applies CoPHo first and DiGress second, \emph{CoPHo*DiGress} applies both controllers at every step.  Table~\ref{tab:qm9} reports the MAE for each target and the average validity rate.

\begin{table}[htbp]
  \centering
  \caption{QM9 molecule generation conditioned on HOMO and \(\mu\).}
  \label{tab:qm9}
  \begin{tabular}{lcccc}
    \toprule
    Model            & HOMO  & \(\mu\) & HOMO+\(\mu\) & Validity(\%) \\
    \midrule
    Uncondition    & 0.93  & 1.71    & 1.34         & 95.4      \\
    DiGress          & 0.56  & \textbf{0.81}    & 0.87         & 96.8      \\
    CoPHo            & 0.64  & 1.24    & 1.07         & \cellcolor{gray!20}97.2      \\
    DiGress+CoPHo    & 0.64  & 1.44    & 0.94         & \textbf{97.6}      \\
    CoPHo+DiGress    & \cellcolor{gray!20}0.49  & \cellcolor{gray!20}0.84    & \textbf{0.83}         & 96.6      \\
    CoPHo*DiGress    & \textbf{0.48}  & 0.93    & \cellcolor{gray!20}0.86         & 95.2      \\
    \bottomrule
  \end{tabular}
\end{table}

\paragraph{Training Time}
We compare the training time required to support \(k\) properties when using DiGress as the diffusion backbone.  Let \(T_d\) denote the time to train the diffusion model and \(T_c\) the time to train a classifier or regressor.  Classifier-guided methods thus require \(T_d + k\,T_c\), whereas other methods require \(k\,T_d\). Table~\ref{tab:training_time} reports \(T_d\) and \(T_c\) for baselines on Enzymes and Community-Small.

\begin{table}[htbp]
  \centering
  \caption{Training time for diffusion backbone ($T_d$) and classifier/regressor ($T_c$) across models. h: hour(s), m: minute(s)}
  \label{tab:training_time}
  {\small
  \begin{tabular}{lcccccccc}
    \toprule
    Dataset   & \multicolumn{2}{c}{Twigs} & \multicolumn{2}{c}{GDSS} & \multicolumn{2}{c}{DiGress} & \multicolumn{2}{c}{CoPHo} \\
              & $T_d$   & $T_c$   & $T_d$   & $T_c$   & $T_d$     & $T_c$     & $T_d$     & $T_c$     \\
    \midrule
    Enzymes   & 6.75h    & --      & 6.75h    & --      & 6.70h      & 10m      & 6.72h      & 03m      \\
    Community & 22m    & --      & 19m    & --      & 20m      & 04m      & 20m      & 01m      \\
    \bottomrule
  \end{tabular}
  }
\end{table}

\paragraph{Cross‐Paradigm Transfer to an SDE‐Based Diffusion Model}

In Tables~\ref{tab:global_single} and~\ref{tab:multi_control}, CoPHo is evaluated on the DDPM‐based DiGress backbone: persistent homology with $T_{\mathrm{homo}}=5$ is applied at every reverse diffusion step. By contrast, Tables~\ref{tab:sde_multi} and~\ref{tab:sde_single} report CoPHo’s performance when transferred to the SDE‐based GDSS model, where we inject persistent homology with $T_{\mathrm{homo}}=1$ only once every ten steps (i.e.\ 2\% of the frequency used in DiGress). Despite this minimal intervention, GDSS+CoPHo significantly outperforms the DDPM‐based variants. This improvement arises because GDSS perturbs node and edge features with continuous Gaussian noise and explicitly constructs a vector field governing distributional transport; within this SDE framework, even sparse classifier‐gradient drift terms can effectively steer samples into the desired region. In contrast, DiGress’s DDPM formulation models discrete probability transitions over nodes and edges without an explicit vector field, which limits the influence of gradient‐based corrections. These results underscore CoPHo’s strong potential to enhance conditional control across diverse diffusion paradigms. 

\begin{table}[htbp]
    \centering
    \caption{Single-global-property conditioning on GDSS}
    \label{tab:sde_single}
    \begin{tabular}{lccc}
      \toprule
      model & clustering & assortativity & transitivity \\
      \midrule
      GDSS \cite{Jo2022}         & 12.1\textcolor{gray}{\scriptsize$\pm$1.6} & 19.6\textcolor{gray}{\scriptsize$\pm$3.9} & 11.4\textcolor{gray}{\scriptsize$\pm$1.6} \\
      GDSS-CoPHo    & \textbf{9.16}\textcolor{gray}{\scriptsize$\pm$0.9} & \textbf{3.58}\textcolor{gray}{\scriptsize$\pm$0.7} & \textbf{1.72}\textcolor{gray}{\scriptsize$\pm$0.4} \\
      \bottomrule
    \end{tabular}
\end{table}

\begin{table}[htbp]
  \centering
    \centering
    \caption{Multi-global-property conditioning on GDSS}
    \label{tab:sde_multi}
    \begin{tabular}{lccc}
      \toprule
      model & clustering & assortativity & transitivity \\
      \midrule
      GDSS \cite{Jo2022}          & 9.45\textcolor{gray}{\scriptsize$\pm$1.4}  & 16.92\textcolor{gray}{\scriptsize$\pm$3.1} & 9.12\textcolor{gray}{\scriptsize$\pm$1.7}  \\
      GDSS-CoPHo    & \textbf{7.94}\textcolor{gray}{\scriptsize$\pm$0.5}  & \textbf{5.27}\textcolor{gray}{\scriptsize$\pm$0.6}  & \textbf{6.93}\textcolor{gray}{\scriptsize$\pm$0.9}  \\
      \bottomrule
    \end{tabular}
\end{table}

\section{Discussion, Conclusion and Limitations}
\label{conclus}
\paragraph{Discussion}
For graphs with \(N\) nodes and \(C\) categories, the classifier gradient with respect to edges or nodes is a tensor of shape \(N \times N \times C\).  Let \(\nabla \mathbf G_{t}[i,j,c]\) denote the gradient at step \(t\) for the potential edge (or node pair) \((i,j)\) and category \(c\).  To compute the \emph{transition tendency} toward a target category \(c^*\), we extract the gradient slice
\[
g^*(i,j) \;=\; \nabla \mathbf G_{t}[i,j,c^*],
\]
and subtract the minimum gradient over all other categories:
\[
\tau(i,j)
= g^*(i,j)\;-\;\min_{c\neq c^*}\,\nabla \mathbf G_{t}[i,j,c].
\]
A larger \(\tau(i,j)\) indicates a stronger inclination to assign category \(c^*\) to \((i,j)\).  We then rank all \((i,j)\) pairs by \(\tau(i,j)\) and apply the corresponding positive or negative update to the graph topology or node labels.  

We implemented this scheme on QM9—treating atom and bond types as categories—and achieved superior performance over prior classifier‐based conditional generation models, validating the efficacy of gradient‐difference conditioning.

For fully connected GNN architectures where every node pair $(i,j)$ is assumed to be connected, CoPHo adapts by interpreting the classifier gradient tensor $\nabla \mathbf G_t[i,j]$ (shape $N \times N$) as both a \emph{tendency} and an \emph{action} signal.  At each denoising step:
\[
\tau(i,j) \;=\; \bigl|\nabla \mathbf  G_t[i,j]\bigr|,
\qquad
s(i,j) \;=\; \operatorname{sign}\!\bigl(\nabla \mathbf G_t[i,j]\bigr).
\]
Here \(\tau(i,j)\) ranks edges by the magnitude of their influence on the target property, and \(s(i,j)\in\{+1,-1\}\) indicates whether to remove (\(+1\)) or add (\(-1\)) the edge.  We then:
\begin{enumerate}
  \item Sort all pairs \((i,j)\) in descending order of \(\tau(i,j)\).
  \item For the top-\(k\) pairs, apply the update \(E_{ij}\gets E_{ij} - s(i,j)\), clipping to \(\{0,1\}\) to maintain binary adjacency (Similar to gradient descent).
  \item Proceed with the next diffusion reverse step on the updated graph.
\end{enumerate}
This procedure leverages the dense initial connectivity to flexibly sculpt the graph toward the desired property, without any retraining of the underlying backbone.

\paragraph{Conclusion}
We have introduced CoPHo, a novel framework for conditional graph diffusion that combines classifier‐gradient guidance with persistent homology filtrations.  By proving that classifier gradients implement the correct density‐ratio reweighting and embedding them into a decreasing filtration of the graph, CoPHo achieves precise conditioning over both global and fine‐grained properties without retraining the diffusion backbone.  Extensive experiments on synthetic and real‐world graph benchmarks, as well as transfer to molecular generation on QM9, demonstrate that CoPHo substantially improves conditional accuracy while preserving or even enhancing sample quality and maintaining competitive efficiency.

\paragraph{Limitations}
\textbf{\textit{Scalability with Shortest‐Path Constraints.}}
While CoPHo scales gracefully to many graph sizes and conditions, its reliance on multiple shortest‐path constraints can become burdensome as the number of conditioned pairs grows.  Conditioning on an increasing fraction of node‐pairs requires constructing and filtering ever‐larger subgraphs, and may demand more sophisticated combinatorial or graph-search algorithms to maintain fidelity.  In practice, enforcing many shortest‐path conditions in a single run leads to a trade-off: stronger conditioning accuracy but significantly slower inference.  Future work must explore adaptive strategies to prune or group constraints intelligently, and to accelerate the persistent homology steps, in order to sustain efficiency in scenarios with extensive fine-grained requirements.  

\textbf{\textit{Fixed Homology Steps and PH Introduction Timing.}}
In the current design, the number of homology steps \(T_{\mathrm{homo}}\) and the PH introduction timing are selected a priori and not adapted during sampling. This scheduling may miss opportunities for more efficient or accurate conditioning. Future work could explore reinforcement learning or other adaptive strategies to jointly decide \(T_{\mathrm{homo}}\) and PH timing on the fly, balancing inference speed and conditioning fidelity. 
\section*{ACKNOWLEDGMENTS}
This work is supported by the National Key R\&D Program of China under Grant
2023YFB2904100.
\bibliographystyle{ACM-Reference-Format}
\bibliography{sample-base}

@String{Computing = "Computing" }

@String{Computer = "{IEEE} Computer" }

@String{Springer = "Springer-Verlag" }

@inproceedings{Zomorodian2005,
  title={Computing persistent homology},
  author={Zomorodian, Afra and Carlsson, Gunnar},
  booktitle={Proceedings of the twentieth annual symposium on Computational geometry},
  pages={347--356},
  year={2004}
}

@article{Dhariwal2021DiffusionBeatGANs,
  title={Diffusion models beat gans on image synthesis},
  author={Dhariwal, Prafulla and Nichol, Alexander},
  journal={Advances in neural information processing systems},
  volume={34},
  pages={8780--8794},
  year={2021}
}

@article{Edelsbrunner2008PH,
  title={Persistent homology-a survey},
  author={Edelsbrunner, Herbert and Harer, John and others},
  journal={Contemporary mathematics},
  volume={453},
  number={26},
  pages={257--282},
  year={2008},
  publisher={Citeseer}
}

@article{Song2019Score,
  title={Generative modeling by estimating gradients of the data distribution},
  author={Song, Yang and Ermon, Stefano},
  journal={Advances in neural information processing systems},
  volume={32},
  year={2019}
}

@article{Ho2020DDPM,
  title={Denoising diffusion probabilistic models},
  author={Ho, Jonathan and Jain, Ajay and Abbeel, Pieter},
  journal={Advances in neural information processing systems},
  volume={33},
  pages={6840--6851},
  year={2020}
}

@article{Song2020ScoreSDE,
  title={Score-based generative modeling through stochastic differential equations},
  author={Song, Yang and Sohl-Dickstein, Jascha and Kingma, Diederik P and Kumar, Abhishek and Ermon, Stefano and Poole, Ben},
  journal={arXiv preprint arXiv:2011.13456},
  year={2020}
}

@article{Qin2023SparseDiff,
  title={Sparse Training of Discrete Diffusion Models for Graph Generation},
  author={Qin, Yiming and Vignac, Cl\'ement and Frossard, Pascal},
  journal={arXiv preprint arXiv:2311.02142},
  year={2023}
}

@article{Sen2008Collective,
  title        = {Collective Classification in Network Data},
  author       = {Sen, Prithviraj and Namata, Galileo and Bilgic, Mustafa and Getoor, Lise and Gallagher, Brian and Eliassi‐Rad, Tina},
  journal      = {AI Magazine},
  volume       = {29},
  number       = {3},
  pages        = {93--106},
  year         = {2008}
}

@article{Schomburg2004BRENDA,
  title        = {BRENDA, the enzyme database: updates and major new developments},
  author       = {Schomburg, Ida and Chang, Antje and Ebeling, Christian and Gremse, Marion and Heldt, Christian and Huhn, Gregor and Schomburg, Dietmar},
  journal      = {Nucleic Acids Research},
  volume       = {32},
  number       = {suppl\_1},
  pages        = {D431--D433},
  year         = {2004},
  doi          = {10.1093/nar/gkh081}
}

@inproceedings{Martinkus2022SPECTRE,
  title        = {SPECTRE: Spectral conditioning helps to overcome the expressivity limits of one-shot graph generators},
  author       = {Martinkus, Karolis and Loukas, Andreas and Perraudin, Nathana{\"e}l and Wattenhofer, Roger},
  booktitle    = {Proceedings of the 39th International Conference on Machine Learning},
  series       = {Proceedings of Machine Learning Research},
  volume       = {162},
  pages        = {15159--15179},
  year         = {2022},
  publisher    = {PMLR}
}

@article{Knight2011TopologyZoo,
  title        = {The Internet Topology Zoo},
  author       = {Knight, Simon and Nguyen, Hung X. and Falkner, Nick and Bowden, Rhys and Roughan, Matthew},
  journal      = {IEEE Journal on Selected Areas in Communications},
  volume       = {29},
  number       = {9},
  pages        = {1765--1775},
  year         = {2011},
  doi          = {10.1109/JSAC.2011.111002}
}

@article{Ramakrishnan2014QM9,
  title={Quantum chemistry structures and properties of 134 kilo molecules},
  author={Ramakrishnan, Raghunathan and Dral, Pavlo O and Rupp, Matthias and Von Lilienfeld, O Anatole},
  journal={Scientific data},
  volume={1},
  number={1},
  pages={1--7},
  year={2014},
  publisher={Nature Publishing Group}
}

@article{Vignac2023DiGress,
  title        = {DiGress: Discrete Denoising Diffusion for Graph Generation},
  author       = {Vignac, Clara and Krawczuk, Igor and Siraudin, Antoine and Wang, Bohan and Cevher, Volkan and Frossard, Pascal},
  journal      = {arXiv preprint arXiv:2209.14734},
  year         = {2023}
}

@article{Bai2019,
  author = {Bai, Ya-Nan and Huang, Ning and Sun, Lina and Wang, Lei},
  title = {Reliability-based topology design for large-scale networks},
  journal = {ISA Transactions},
  volume = {94},
  pages = {144--150},
  year = {2019}
}

@inproceedings{You2018,
  title={Graphrnn: Generating realistic graphs with deep auto-regressive models},
  author={You, Jiaxuan and Ying, Rex and Ren, Xiang and Hamilton, William and Leskovec, Jure},
  booktitle={International conference on machine learning},
  pages={5708--5717},
  year={2018},
  organization={PMLR}
}

@inproceedings{Simonovsky2018,
  title={Graphvae: Towards generation of small graphs using variational autoencoders},
  author={Simonovsky, Martin and Komodakis, Nikos},
  booktitle={Artificial Neural Networks and Machine Learning--ICANN 2018: 27th International Conference on Artificial Neural Networks, Rhodes, Greece, October 4-7, 2018, Proceedings, Part I 27},
  pages={412--422},
  year={2018},
  organization={Springer}
}

@article{DeCao2018,
  title={MolGAN: An implicit generative model for small molecular graphs},
  author={De Cao, Nicola and Kipf, Thomas},
  journal={arXiv preprint arXiv:1805.11973},
  year={2018}
}

@inproceedings{Jo2022,
  author = {Jo, Jaehyeong and Lee, Seul and Hwang, Sung Ju},
  title = {{Score-based Generative Modeling of Graphs via the System of Stochastic Differential Equations}},
  booktitle = {Proceedings of the 39th International Conference on Machine Learning (ICML)},
  pages = {10362--10383},
  year = {2022}
}

@article{Mercatali2024,
  title={Diffusion twigs with loop guidance for conditional graph generation},
  author={Mercatali, Giangiacomo and Verma, Yogesh and Freitas, Andre and Garg, Vikas},
  journal={Advances in Neural Information Processing Systems},
  volume={37},
  pages={137741--137767},
  year={2024}
}

@inproceedings{Lee2023,
  title={Exploring chemical space with score-based out-of-distribution generation},
  author={Lee, Seul and Jo, Jaehyeong and Hwang, Sung Ju},
  booktitle={International Conference on Machine Learning},
  pages={18872--18892},
  year={2023},
  organization={PMLR}
}

@article{Edelsbrunner2002,
  author = {Edelsbrunner, Herbert and Letscher, David and Zomorodian, Afra},
  title = {{Topological Persistence and Simplification}},
  journal = {Discrete \& Computational Geometry},
  volume = {28},
  number = {4},
  pages = {511--533},
  year = {2002}
}

@inproceedings{hoogeboom2022equivariant,
  title={Equivariant diffusion for molecule generation in 3d},
  author={Hoogeboom, Emiel and Satorras, V{\i}ctor Garcia and Vignac, Cl{\'e}ment and Welling, Max},
  booktitle={International conference on machine learning},
  pages={8867--8887},
  year={2022},
  organization={PMLR}
}

@article{huang2024learning,
  title={Learning joint 2-d and 3-d graph diffusion models for complete molecule generation},
  author={Huang, Han and Sun, Leilei and Du, Bowen and Lv, Weifeng},
  journal={IEEE Transactions on Neural Networks and Learning Systems},
  year={2024},
  publisher={IEEE}
}

@inproceedings{xu2023geometric,
  title={Geometric latent diffusion models for 3d molecule generation},
  author={Xu, Minkai and Powers, Alexander S and Dror, Ron O and Ermon, Stefano and Leskovec, Jure},
  booktitle={International Conference on Machine Learning},
  pages={38592--38610},
  year={2023},
  organization={PMLR}
}

@article{xu2024discrete,
  title={Discrete-state Continuous-time Diffusion for Graph Generation},
  author={Xu, Zhe and Qiu, Ruizhong and Chen, Yuzhong and Chen, Huiyuan and Fan, Xiran and Pan, Menghai and Zeng, Zhichen and Das, Mahashweta and Tong, Hanghang},
  journal={arXiv preprint arXiv:2405.11416},
  year={2024}
}

@article{bao2022equivariant,
  title={Equivariant energy-guided sde for inverse molecular design},
  author={Bao, Fan and Zhao, Min and Hao, Zhongkai and Li, Peiyao and Li, Chongxuan and Zhu, Jun},
  journal={arXiv preprint arXiv:2209.15408},
  year={2022}
}

@inproceedings{kim2022equalnet,
  title={EqualNet: A Secure and Practical Defense for Long-term Network Topology Obfuscation.},
  author={Kim, Jinwoo and Marin, Eduard and Conti, Mauro and Shin, Seungwon},
  booktitle={NDSS},
  year={2022}
}

@techreport{hagberg2008exploring,
  title={Exploring network structure, dynamics, and function using NetworkX},
  author={Hagberg, Aric and Swart, Pieter J and Schult, Daniel A},
  year={2008},
  institution={Los Alamos National Laboratory (LANL), Los Alamos, NM (United States)}
}

@article{yang2020secure,
  title={Secure deep graph generation with link differential privacy},
  author={Yang, Carl and Wang, Haonan and Zhang, Ke and Chen, Liang and Sun, Lichao},
  journal={arXiv preprint arXiv:2005.00455},
  year={2020}
}

@inproceedings{yuan2023privgraph,
  title={$\{$PrivGraph$\}$: differentially private graph data publication by exploiting community information},
  author={Yuan, Quan and Zhang, Zhikun and Du, Linkang and Chen, Min and Cheng, Peng and Sun, Mingyang},
  booktitle={32nd USENIX Security Symposium (USENIX Security 23)},
  pages={3241--3258},
  year={2023}
}

@inproceedings{leskovec2010signed,
  title={Signed networks in social media},
  author={Leskovec, Jure and Huttenlocher, Daniel and Kleinberg, Jon},
  booktitle={Proceedings of the SIGCHI conference on human factors in computing systems},
  pages={1361--1370},
  year={2010}
}

@article{leskovec2007graph,
  title={Graph evolution: Densification and shrinking diameters},
  author={Leskovec, Jure and Kleinberg, Jon and Faloutsos, Christos},
  journal={ACM transactions on Knowledge Discovery from Data (TKDD)},
  volume={1},
  number={1},
  pages={2--es},
  year={2007},
  publisher={ACM New York, NY, USA}
}

@article{qian2017social,
  title={Social network de-anonymization and privacy inference with knowledge graph model},
  author={Qian, Jianwei and Li, Xiang-Yang and Zhang, Chunhong and Chen, Linlin and Jung, Taeho and Han, Junze},
  journal={IEEE Transactions on Dependable and Secure Computing},
  volume={16},
  number={4},
  pages={679--692},
  year={2017},
  publisher={IEEE}
}

@article{alrumaih2023genind,
  title={GENIND: An industrial network topology generator},
  author={Alrumaih, Thuraya NI and Alenazi, Mohammed JF},
  journal={Alexandria Engineering Journal},
  volume={79},
  pages={56--71},
  year={2023},
  publisher={Elsevier}
}

@article{dadauto2024data,
  title={Dadauto, Caio V},
  author={Dadauto, Caio V and da Fonseca, Nelson LS and Torres, Ricardo da S},
  journal={IEEE Transactions on Network and Service Management},
  year={2024},
  publisher={IEEE}
}

@article{tangmunarunkit2002network,
  title={Network topology generators: Degree-based vs. structural},
  author={Tangmunarunkit, Hongsuda and Govindan, Ramesh and Jamin, Sugih and Shenker, Scott and Willinger, Walter},
  journal={ACM SIGCOMM Computer Communication Review},
  volume={32},
  number={4},
  pages={147--159},
  year={2002},
  publisher={ACM New York, NY, USA}
}

@inproceedings{sharafaldin2019developing,
  title={Developing realistic distributed denial of service (DDoS) attack dataset and taxonomy},
  author={Sharafaldin, Iman and Lashkari, Arash Habibi and Hakak, Saqib and Ghorbani, Ali A},
  booktitle={2019 international carnahan conference on security technology (ICCST)},
  pages={1--8},
  year={2019},
  organization={IEEE}
}

@article{molnar2023threshold,
  title={Threshold sensitivity of the production network topology},
  author={Moln{\'a}r, Eszter and Csala, D{\'e}nes},
  journal={Applied Network Science},
  volume={8},
  number={1},
  pages={71},
  year={2023},
  publisher={Springer}
}

@article{cavallaro2024sensitivity,
  title={On the sensitivity of centrality metrics},
  author={Cavallaro, Lucia and De Meo, Pasquale and Fiumara, Giacomo and Liotta, Antonio},
  journal={Plos one},
  volume={19},
  number={5},
  pages={e0299255},
  year={2024},
  publisher={Public Library of Science San Francisco, CA USA}
}

@article{kipf2016semi,
  title={Semi-supervised classification with graph convolutional networks},
  author={Kipf, Thomas N and Welling, Max},
  journal={arXiv preprint arXiv:1609.02907},
  year={2016}
}

@article{hamilton2017inductive,
  title={Inductive representation learning on large graphs},
  author={Hamilton, Will and Ying, Zhitao and Leskovec, Jure},
  journal={Advances in neural information processing systems},
  volume={30},
  year={2017}
}

@article{xu2018powerful,
  title={How powerful are graph neural networks?},
  author={Xu, Keyulu and Hu, Weihua and Leskovec, Jure and Jegelka, Stefanie},
  journal={arXiv preprint arXiv:1810.00826},
  year={2018}
}

@article{kreuzer2021rethinking,
  title={Rethinking graph transformers with spectral attention},
  author={Kreuzer, Devin and Beaini, Dominique and Hamilton, Will and L{\'e}tourneau, Vincent and Tossou, Prudencio},
  journal={Advances in Neural Information Processing Systems},
  volume={34},
  pages={21618--21629},
  year={2021}
}

@article{yun2019graph,
  title={Graph transformer networks},
  author={Yun, Seongjun and Jeong, Minbyul and Kim, Raehyun and Kang, Jaewoo and Kim, Hyunwoo J},
  journal={Advances in neural information processing systems},
  volume={32},
  year={2019}
}

@article{ying2021transformers,
  title={Do transformers really perform badly for graph representation?},
  author={Ying, Chengxuan and Cai, Tianle and Luo, Shengjie and Zheng, Shuxin and Ke, Guolin and He, Di and Shen, Yanming and Liu, Tie-Yan},
  journal={Advances in neural information processing systems},
  volume={34},
  pages={28877--28888},
  year={2021}
}

@inproceedings{kong2023goat,
  title={GOAT: A global transformer on large-scale graphs},
  author={Kong, Kezhi and Chen, Jiuhai and Kirchenbauer, John and Ni, Renkun and Bruss, C Bayan and Goldstein, Tom},
  booktitle={International Conference on Machine Learning},
  pages={17375--17390},
  year={2023},
  organization={PMLR}
}

@inproceedings{melancon2006just,
  title={Just how dense are dense graphs in the real world? A methodological note},
  author={Melancon, Guy},
  booktitle={Proceedings of the 2006 AVI workshop on BEyond time and errors: novel evaluation methods for information visualization},
  pages={1--7},
  year={2006}
}

@article{casiraghi2025empirical,
  title={Empirical networks are sparse: Enhancing multiedge models with zero-inflation},
  author={Casiraghi, Giona and Andres, Georges},
  journal={PNAS nexus},
  volume={4},
  number={1},
  pages={pgaf001},
  year={2025},
  publisher={Oxford University Press US}
}

@inproceedings{hofer2020graph,
  title={Graph filtration learning},
  author={Hofer, Christoph and Graf, Florian and Rieck, Bastian and Niethammer, Marc and Kwitt, Roland},
  booktitle={International Conference on Machine Learning},
  pages={4314--4323},
  year={2020},
  organization={PMLR}
}

@article{nguyen2025persistent,
  title={Persistent Homology-induced Graph Ensembles for Time Series Regressions},
  author={Nguyen, Viet The and Pham, Duy Anh and Le, An Thai and Peter, Jans and Gust, Gunther},
  journal={arXiv preprint arXiv:2503.14240},
  year={2025}
}

@article{yan2025enhancing,
  title={Enhancing Graph Representation Learning with Localized Topological Features},
  author={Yan, Zuoyu and Zhao, Qi and Ye, Ze and Ma, Tengfei and Gao, Liangcai and Tang, Zhi and Wang, Yusu and Chen, Chao},
  journal={Journal of Machine Learning Research},
  volume={26},
  number={5},
  pages={1--36},
  year={2025}
}

@inproceedings{rombach2022latent,
  title     = {High‐Resolution Image Synthesis with Latent Diffusion Models},
  author    = {Rombach, Robin and Blattmann, Andreas and Lorenz, Dominik and Esser, Patrick and Ommer, Björn},
  booktitle = {IEEE/CVF Conference on Computer Vision and Pattern Recognition},
  pages     = {10684--10695},
  year      = {2022},
  url       = {https://arxiv.org/abs/2112.10752}
}

@inproceedings{kong2020diffwave,
  title     = {DiffWave: A Versatile Diffusion Model for Audio Synthesis},
  author    = {Kong, Zhifeng and Ping, Wei and Huang, Jiaji and Zhao, Kexin and Catanzaro, Bryan},
  booktitle = {International Conference on Learning Representations},
  year      = {2021},
  url       = {https://arxiv.org/abs/2009.09761}
}

@inproceedings{popov2021gradtts,
  title     = {Grad‐TTS: A Diffusion Probabilistic Model for Text‐to‐Speech},
  author    = {Popov, Ruslan and Wang, Yannis and Lijiang, Yu and Lei, Mai and Lee, Zheng and Xu, Yi and Weng, Shuang and Liu, Zhe and Foster, Guy and Rudnicky, Alexander and Zen, Hisashi},
  booktitle = {Interspeech},
  pages     = {3645--3649},
  year      = {2021},
  url       = {https://arxiv.org/abs/2105.01601}
}

@article{chen2023efficient,
  title={Efficient and degree-guided graph generation via discrete diffusion modeling},
  author={Chen, Xiaohui and He, Jiaxing and Han, Xu and Liu, Li-Ping},
  journal={arXiv preprint arXiv:2305.04111},
  year={2023}
}

@article{gebauer2019symmetry,
  title={Symmetry-adapted generation of 3d point sets for the targeted discovery of molecules},
  author={Gebauer, Niklas and Gastegger, Michael and Sch{\"u}tt, Kristof},
  journal={Advances in neural information processing systems},
  volume={32},
  year={2019}
}

@article{gebauer2022inverse,
  title={Inverse design of 3d molecular structures with conditional generative neural networks},
  author={Gebauer, Niklas WA and Gastegger, Michael and Hessmann, Stefaan SP and M{\"u}ller, Klaus-Robert and Sch{\"u}tt, Kristof T},
  journal={Nature communications},
  volume={13},
  number={1},
  pages={973},
  year={2022},
  publisher={Nature Publishing Group UK London}
}

@article{geng2023novo,
  title={De novo molecular generation via connection-aware motif mining},
  author={Geng, Zijie and Xie, Shufang and Xia, Yingce and Wu, Lijun and Qin, Tao and Wang, Jie and Zhang, Yongdong and Wu, Feng and Liu, Tie-Yan},
  journal={arXiv preprint arXiv:2302.01129},
  year={2023}
}

@article{gruver2023protein,
  title={Protein design with guided discrete diffusion},
  author={Gruver, Nate and Stanton, Samuel and Frey, Nathan and Rudner, Tim GJ and Hotzel, Isidro and Lafrance-Vanasse, Julien and Rajpal, Arvind and Cho, Kyunghyun and Wilson, Andrew G},
  journal={Advances in neural information processing systems},
  volume={36},
  pages={12489--12517},
  year={2023}
}

@article{li2023graphmaker,
  title={Graphmaker: Can diffusion models generate large attributed graphs?},
  author={Li, Mufei and Krea{\v{c}}i{\'c}, Eleonora and Potluru, Vamsi K and Li, Pan},
  journal={arXiv preprint arXiv:2310.13833},
  year={2023}
}

@article{nisonoff2024unlocking,
  title={Unlocking guidance for discrete state-space diffusion and flow models},
  author={Nisonoff, Hunter and Xiong, Junhao and Allenspach, Stephan and Listgarten, Jennifer},
  journal={arXiv preprint arXiv:2406.01572},
  year={2024}
}

@inproceedings{jiang2024diffkg,
  title={Diffkg: Knowledge graph diffusion model for recommendation},
  author={Jiang, Yangqin and Yang, Yuhao and Xia, Lianghao and Huang, Chao},
  booktitle={Proceedings of the 17th ACM international conference on web search and data mining},
  pages={313--321},
  year={2024}
}

@article{liu2024graph,
  title={Graph diffusion policy optimization},
  author={Liu, Yijing and Du, Chao and Pang, Tianyu and Li, Chongxuan and Lin, Min and Chen, Wei},
  journal={Advances in Neural Information Processing Systems},
  volume={37},
  pages={9585--9611},
  year={2024}
}

@inproceedings{you2024latent,
  title={Latent 3d graph diffusion},
  author={You, Yuning and Zhou, Ruida and Park, Jiwoong and Xu, Haotian and Tian, Chao and Wang, Zhangyang and Shen, Yang},
  year={2024},
  organization={International Conference on Learning Representations (ICLR)}
}

@article{liu2024graphdif,
  title={Graph diffusion transformers for multi-conditional molecular generation},
  author={Liu, Gang and Xu, Jiaxin and Luo, Tengfei and Jiang, Meng},
  journal={arXiv preprint arXiv:2401.13858},
  year={2024}
}

@article{erdos1960evolution,
  title={On the evolution of random graphs},
  author={Erdos, Paul and R{\'e}nyi, Alfr{\'e}d and others},
  journal={Publ. math. inst. hung. acad. sci},
  volume={5},
  number={1},
  pages={17--60},
  year={1960}
}

@article{Dijkstra1959,
  author  = {Dijkstra, E. W.},
  title   = {A Note on Two Problems in Connexion with Graphs},
  journal = {Numerische Mathematik},
  volume  = {1},
  number  = {1},
  pages   = {269--271},
  year    = {1959},
  doi     = {10.1007/BF01386390},
}

@inproceedings{sharma2024diffuse,
  title={Diffuse, sample, project: plug-and-play controllable graph generation},
  author={Sharma, Kartik and Kumar, Srijan and Trivedi, Rakshit},
  booktitle={Forty-first International Conference on Machine Learning},
  year={2024}
}

@inproceedings{gutfraind2015multiscale,
  title={Multiscale network generation},
  author={Gutfraind, Alexander and Safro, Ilya and Meyers, Lauren Ancel},
  booktitle={2015 18th international conference on information fusion (fusion)},
  pages={158--165},
  year={2015},
  organization={IEEE}
}

@article{ermagun2021recovery,
  title={Recovery patterns and physics of the network},
  author={Ermagun, Alireza and Tajik, Nazanin},
  journal={PloS one},
  volume={16},
  number={1},
  pages={e0245396},
  year={2021},
  publisher={Public Library of Science San Francisco, CA USA}
}

@inproceedings{patro2012missing,
  title={The missing models: A data-driven approach for learning how networks grow},
  author={Patro, Robert and Duggal, Geet and Sefer, Emre and Wang, Hao and Filippova, Darya and Kingsford, Carl},
  booktitle={Proceedings of the 18th ACM SIGKDD international conference on Knowledge discovery and data mining},
  pages={42--50},
  year={2012}
}

\appendix

\section{Addition Results}
\label{sec:addres}
\subsection{More ablations}
\label{more_abla}
\paragraph{Persistent Homology}
Conditional control without persistent homology is practically equivalent to DiGress \cite{Vignac2023DiGress}.  We evaluate the impact of adding persistent homology on both global and fine‐grained conditional control tasks using Community-Small and Planar.  As shown in Table~\ref{tab:ablation_homology}, introducing persistent homology yields substantial improvements in mean absolute error for density, clustering, assortativity, transitivity and shortest‐path condition across all tasks.

\begin{table}[htbp]
  \centering
  \caption{Ablation Study on Persistent Homology. D: Density, C: Clustering, A: Assortativity, T: Transitivity. w/o: without.}
  \label{tab:ablation_homology}
  \begin{tabular}{lcc|cccc}
    \toprule
    & \multicolumn{2}{c}{Planar} & \multicolumn{4}{c}{Community-small} \\
    Configuration               & MAE  & KL    & D & C & A & T \\
    \midrule
    w/o PH & 1.15 & 0.09  & 2.19  & 9.56     & 16.9        & 8.72       \\
    w/ PH    & 0.43 & 0.02  & 1.89  & 7.13     & 10.9       & 7.23       \\
    \bottomrule
  \end{tabular}
\end{table}

\subsection{Generation Quality Under Property Conditioning}
\label{gene_quality}
We also generation quality under global property conditioning and report average performance across all conditional targets. Table~\ref{tab:gen_quality_global} presents the mean values of the three metrics described in Sec.~\ref{sec:setup}.

\begin{table}[ht]
  \centering
  \caption{Generation quality metrics under property conditioning.}
  \label{tab:gen_quality_global}
  \begin{tabular}{lccc|ccc}
    \toprule
     & \multicolumn{3}{c}{Community-Small} & \multicolumn{3}{c}{Enzymes} \\
    Model    & Deg. & Clus. & Orb. & Spec. & Clus. & Orb. \\
    \midrule
    DiGress  & 0.02 & 11.7 & \textbf{1.16} & \textbf{1.87} & \textbf{3.10} & \textbf{1.58} \\
    Ours     & \textbf{0.02} & \textbf{10.3} & 1.17 & 2.59 & 3.32 & 2.07 \\
    \bottomrule
  \end{tabular}
\end{table}

\subsection{Parameterized Posterior Distribution of Discrete Graph Diffusion}
\label{sec:posterior_vis}
\begin{figure}[htbp]        
  \centering               
  \includegraphics[width=0.5\textwidth]{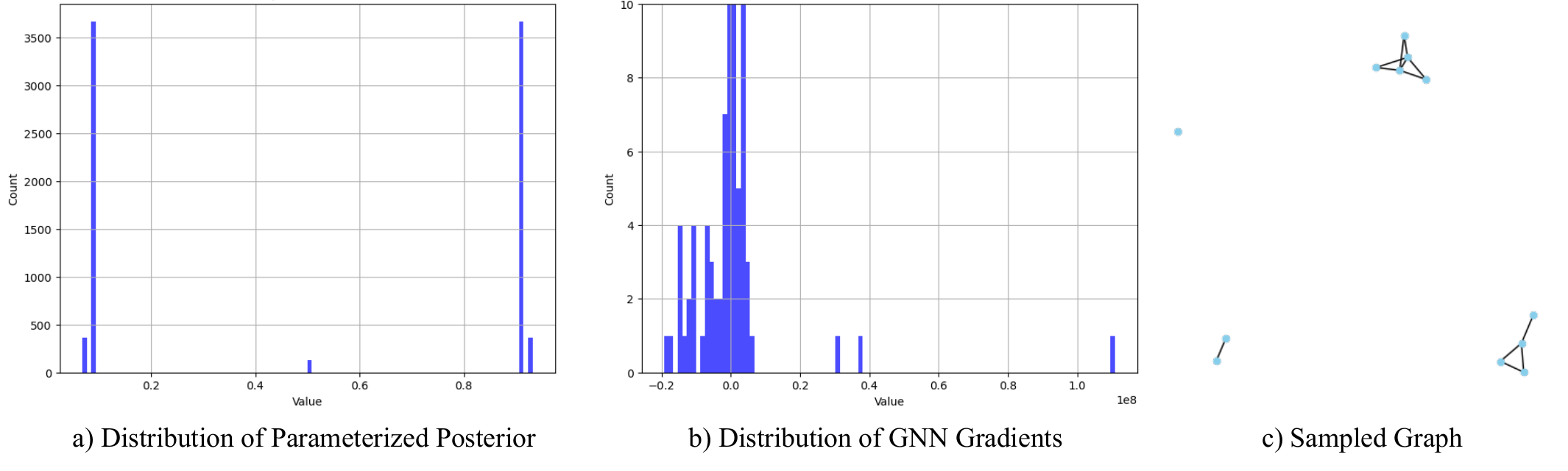}
  \caption{Generate sample resulting from posterior and gradient guidance.}
  \Description{digress fail case}
  \label{fig:badres}
\end{figure}
We visualize the parameterized posterior distribution of the unconditional discrete graph diffusion model, alongside the gradients on the adjacency matrix edges with respect to the shortest‐path property as predicted by a GNN.  Using the DiGress framework on the Planar dataset, we render these in Figure~\ref{fig:badres}.  At every step, the posterior concentrates sharply, while the edge gradients exhibit a very \textbf{peaky} distribution (ranging from \(-0.2\times10^{7}\) to \(1\times10^{8}\), with a single outlier near \(10^{8}\)).  Consequently, when we normalize these gradients and inject them into the posterior, it becomes impossible to choose a guidance strength that preserves the (approximate) continuity assumption on graph data.  After normalization, nearly all guidance scores collapse to zero, causing the sampled graphs to become progressively sparser at each diffusion step.  For example, in the Figure~\ref{fig:badres}-b, only one edge retains a gradient on the order of \(10^{8}\), and the rest are effectively zero—producing the degenerate sample shown in Figure~\ref{fig:badres}-c.

\section{Detailed Derivations and Proofs}
\label{sec:proof_full}
\subsection{Derivation: Sampling Process}
\label{sec:derivation_sampling}

We wish to model the conditional sampling step
\[
\mathbf{G}_{t+1}\;\longrightarrow\;\hat{\mathbf G}_t
\]
by introducing the intermediate (unconditional) denoised graph \(\mathbf G_t\).  We proceed in three steps:

\paragraph{1. Law of Total Probability}  
For any two random variables \(X,Y\), the law of total probability states
\[
\Pr(X\mid Y)
=\sum_{z}\Pr(X,z\mid Y).
\]
Applying this to our chain gives
\[
q(\hat{\mathbf G}_t \mid \mathbf G_{t+1})
=\sum_{\mathbf G_t}q\bigl(\hat{\mathbf G}_t,\mathbf G_t\mid \mathbf G_{t+1}\bigr).
\]

\paragraph{2. Chain Rule (Product Rule)}  
The joint conditional can be factorized via the chain rule:
\[
q\bigl(\hat{\mathbf G}_t,\mathbf G_t\mid \mathbf G_{t+1}\bigr)
= q\bigl(\hat{\mathbf G}_t\mid \mathbf G_t,\mathbf G_{t+1}\bigr)
  \;q\bigl(\mathbf G_t\mid \mathbf G_{t+1}\bigr).
\]

\paragraph{3. Markov Property}  
Under the two‐stage Markov assumption
\(\mathbf G_{t+1}\to \mathbf G_t\to \hat{\mathbf G}_t\),
the intermediate state \(\mathbf G_t\) renders \(\hat{\mathbf G}_t\) conditionally independent of \(\mathbf G_{t+1}\):
\[
q\bigl(\hat{\mathbf G}_t\mid \mathbf G_t,\mathbf G_{t+1}\bigr)
= q\bigl(\hat{\mathbf G}_t\mid \mathbf G_t\bigr).
\]

\paragraph{Combine All Steps}  
Putting (1)–(3) together, we obtain
\[
\begin{aligned}
q(\hat{\mathbf G}_t \mid \mathbf G_{t+1})
&= \sum_{\mathbf G_t}
     q\bigl(\hat{\mathbf G}_t,\mathbf G_t\mid \mathbf G_{t+1}\bigr)\\
&= \sum_{\mathbf G_t}
     q\bigl(\hat{\mathbf G}_t\mid \mathbf G_t,\mathbf G_{t+1}\bigr)
     \;q\bigl(\mathbf G_t\mid \mathbf G_{t+1}\bigr)\\
&= \sum_{\mathbf G_t}
     q\bigl(\hat{\mathbf G}_t\mid \mathbf G_t\bigr)
     \;q\bigl(\mathbf G_t\mid \mathbf G_{t+1}\bigr)\,,
\end{aligned}
\]
which is the desired result.

\subsection{Proof of the Unbiased Importance Weight}
\label{sec:proof_unbiased_weight}
\paragraph{Definition of Expected Value for Discrete Random Variables} For a discrete random variable \(X\) with probability mass function \(p(x)\), the expected value (mean) is defined as the weighted sum of its possible values:
\begin{equation}
\label{eq:defexp}
    \mathbb{E}[X]
=
\sum_{x} x\,p(x). 
\tag{A1}
\end{equation}

This follows directly from the definition of expectation for discrete random variables, interpreting it as a probability-weighted average of outcomes.

Let the true conditional kernel be
\[
q(\mathbf{\hat{G}}_t \mid \mathbf{G}_{t+1}, y)
\;\propto\;
q(\mathbf{G}_t \mid \mathbf{G}_{t+1})\,q(y \mid \mathbf{G}_t),
\]
and let our proposal distribution be
\[
\mathbb{P}(\mathbf{\hat{G}}_t \mid \mathbf{G}_t).
\]
We define the importance weight
\[
w(\mathbf{\hat{G}}_t)
=\frac{q(\mathbf{\hat{G}}_t \mid \mathbf{G}_{t+1})\,q(y \mid \mathbf{\hat{G}}_t)}
      {\mathbb{P}(\mathbf{\hat{G}}_t \mid \mathbf{G}_t)}.
\]
We now show that for any test function \(f\), the weighted estimator is unbiased, as definition in Eq.~\ref{eq:defexp}:
\[
\begin{aligned}
\mathbb{E}_{p}\bigl[w(\mathbf{\hat{G}}_t)\,f(\mathbf{\hat{G}}_t)\bigr]
&=\sum_{\mathbf{\hat{G}}_t} \mathbb{P}(\mathbf{\hat{G}}_t \mid \mathbf{G}_t)\,
  \frac{q(\mathbf{\hat{G}}_t \mid \mathbf{G}_{t+1})\,q(y \mid \mathbf{\hat{G}}_t)}
       {\mathbb{P}(\mathbf{\hat{G}}_t \mid \mathbf{G}_t)}\,
  f(\mathbf{\hat{G}}_t)\\
&=\sum_{\mathbf{\hat{G}}_t}q(\mathbf{\hat{G}}_t \mid \mathbf{G}_{t+1})\,q(y \mid \mathbf{\hat{G}}_t)\,f(\mathbf{\hat{G}}_t)\\
&\propto
\sum_{\mathbf{\hat{G}}_t}q\bigl(\mathbf{\hat{G}}_t \mid \mathbf{G}_{t+1},y\bigr)\,f(\mathbf{\hat{G}}_t)
=\mathbb{E}_{q}\bigl[f(\mathbf{\hat{G}}_t)\bigr].
\end{aligned}
\]
In particular, setting \(f(\mathbf{\hat{G}}_t)\equiv 1\) gives
\[
\mathbb{E}_{p}\bigl[w(\mathbf{\hat{G}}_t)\bigr]
=\sum_{\mathbf{\hat{G}}_t}q(\mathbf{\hat{G}}_t \mid \mathbf{G}_{t+1})\,q(y \mid \mathbf{\hat{G}}_t)
=\sum_{\mathbf{\hat{G}}_t}q\bigl(\mathbf{\hat{G}}_t \mid \mathbf{G}_{t+1},y\bigr)
=1,
\]
showing that \(w(\mathbf{\hat{G}}_t)\) is indeed an unbiased importance weight.

\subsection{Proof of Gradient as Proposal Distribution}
\label{sec:proof_gradient_proposal}

To approximate sampling from the intractable conditional kernel
\(\displaystyle q(\hat{\mathbf G}_t\mid \mathbf G_t,\, y)\),
we use importance sampling.  The optimal proposal is
\[
q^*(\mathbf G_t)\;\propto\;q(\mathbf G_t\mid \mathbf G_{t+1})\;q(y\mid \mathbf G_t)\,.
\]
In continuous diffusion, classifier guidance multiplies the base score by
\(\exp\{\ln q( y\mid \mathbf{G}_t)\}\) (exponential tilting) and is realized by adding
\(\nabla_{G_t}\ln q(y\mid \mathbf{G}_t)\) to the denoising gradient.

In the discrete‐graph case, we approximate this exponential tilt by gradually \textbf{removing} the edge \(e^*\) whose classifier‐gradient
\(\frac{\partial}{\partial A_e}\ln q(y\mid \mathbf{G}_t)\)
are top-k maximal.  Concretely:
\[
e^* \;=\;\arg\max_{e\in E(G_t)}
\partial_{A_e}\ln q(y\mid \mathbf{G}_t)\\, 
\quad
\mathbf{\hat G}_t \;=\; \mathbf{G}_t\setminus\{e^*\}\,.
\]
This construction yields a proposal distribution
\(\displaystyle p(\mathbf{\hat G}_t\mid \mathbf{G}_t)\)
that concentrates mass on high‐tilt moves and thus closely matches \(q^*(\mathbf{G}_t)\), reducing variance in importance‐sampling estimates.


\subsection{Discussion on the conditional markov chain}
\paragraph{1. Memorylessness}  
The reverse diffusion process is naturally formulated as a Markov chain because each state transition depends only on the immediately preceding graph, not on the full history.  This “memoryless” property ensures that the distribution of \(\mathbf G_t\) conditioned on \(\mathbf G_{t+1}\) is independent of all earlier states \(\mathbf G_{t+2}, \mathbf G_{t+3},\ldots\) .

\paragraph{2. Two‐Stage Decomposition}  
Each conditional sampling step is defined as a two‐stage chain: first, an unconditional denoising from \(\mathbf G_{t+1}\) to \(\mathbf G_t\) via the learned kernel \(q(\mathbf G_t\!\mid\!\mathbf G_{t+1})\); second, a conditional refinement to \(\hat{\mathbf G}_t\) that incorporates the target \(y\). So the process can be regarded as a markov chain.

\end{document}